# A Comprehensive Review of Multi-Agent Reinforcement Learning in Video Games

Zhengyang Li , *Member, IEEE*, Qijin Ji , Xinghong Ling , and Quan Liu

*Abstract*— Recent advancements in multi-agent reinforcement learning (MARL) have demonstrated its application potential in modern games. Beginning with foundational work and progressing to landmark achievements such as AlphaStar in *StarCraft II* and OpenAI Five in *Dota 2*, MARL has proven capable of achieving superhuman performance across diverse game environments through techniques like self-play, supervised learning, and deep reinforcement learning. With its growing impact, a comprehensive review has become increasingly important in this field. This paper aims to provide a thorough examination of MARL's application from turn-based two-agent games to real-time multi-agent video games including popular genres such as Sports games, First-Person Shooter (FPS) games, Real-Time Strategy (RTS) games and Multiplayer Online Battle Arena (MOBA) games. We further analyze critical challenges posed by MARL in video games, including nonstationary, partial observability, sparse rewards, team coordination, and scalability, and highlight successful implementations in games like *Rocket League*, *Minecraft*, *Quake III Arena*, *StarCraft II*, *Dota 2*, *Honor of Kings*, etc. This paper offers insights into MARL in video game AI systems, proposes a novel method to estimate game complexity, and suggests future research directions to advance MARL and its applications in game development, inspiring further innovation in this rapidly evolving field.

## I. INTRODUCTION

AS of July 2023, more than 212.6 million people in the U.S. play video games regularly, accounting for approximately two-thirds of its entire population [1]. This significant engagement extends beyond the U.S., with an even more pronounced trend globally, where the number explodes to 3.22 billion [2].

Over the past several decades, video games have undergone a fundamental transformation from primarily single-player or turn-based titles to real-time, multiplayer formats. Today's most popular genres include MOBA, RTS, and FPS, all of which are designed around team-based, competitive, or cooperative multiplayer gameplay. As shown in **Fig. 1**, 88% of players now report having played games online, indicating a dominant trend toward networked, interaction-rich gameplay environments. However, due to limitations such as player

Zhengyang Li is with DigiPen Institute of Technology, Redmond, WA 98052, USA (e-mail: zhengyang.li@digipen.edu).

Qijin Ji (corresponding author) and Xinghong Ling are with the School of Computational Science and Artificial Intelligence, Suzhou City University, Suzhou, Jiangsu 215104, China. (e-mail q_ji, lingxinghong@szcu.edu.cn).

Quan Liu is with the School of Computer Science and Technology, Soochow University, Suzhou, Jiangsu 215006, China (e-mail: quanliu@suda.edu.cn).

## Games Foster Community

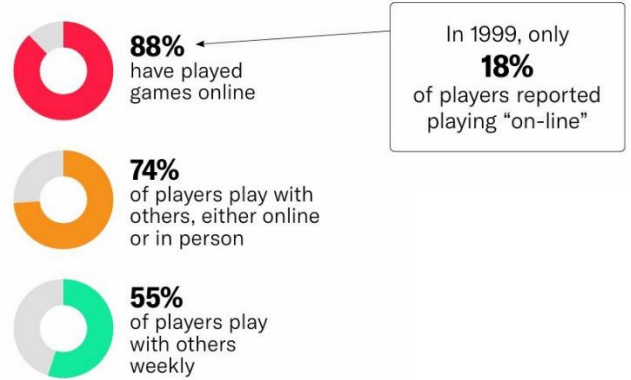

Fig. 1. This figure contrasts the significant growth in online gaming, showing that 88% of players have played games online, up from just 18% in 1999 [3].

dropout, matchmaking delays, game trainings, or single-player modes in multiplayer-centric games, artificial intelligence (AI)-controlled agents are frequently introduced to fill the roles of human participants. This growing reliance on artificial teammates and opponents has accelerated the demand for robust multi-agent AI systems capable of modeling coordination, adversarial behavior, and adaptation to human strategies. Furthermore, genres such as strategy and social simulation games offer ideal testbeds for MARL, as they naturally involve complex, persistent interactions among multiple autonomous agents.

For more than half a century, AI has been a critical part of video games ever since their first application in the 1940s [4]. Early games like *Space Invaders* and *Pac-Man* in the 1970s and 1980s featured basic game AI systems to manage non-player character (NPC) behaviors that provided predictable yet engaging challenges for players. These fundamental AI systems paved the way for more complex and smarter behaviors seen in modern games. Today, in titles such as *StarCraft*, *Age of Empires*, *Sid Meier's Civilization*, *Call of Duty*, *Counter-Strike*, *Grand Theft Auto*, *League of Legends*, *Dota 2*, and many others, AI technologies have evolved to address a large number of challenges with intelligent and efficient planning and decision-making that enhance gaming experience and strategic depth. As game environments have evolved from simple 2D planes 3D worlds, the complexity and realism of these environments have increased exponentially. This evolution has turned video games into a field of study that not only improves AI interactions in those games, but also offers a valuable platform for applying different AI techniques in real-world.

Modern games implement AI as a set of algorithms or systems to manage numerous decisions and planning tasks required in games, such as pathfinding, resource management, combats, player or NPC behaviors, and player interactions. Established techniques, such as finite state machines and behavior trees [5] are frequently used to drive AI in games. However, as games have become increasingly realistic, players have begun to desire more intelligent and more realistic NPCs, whether as cooperative companions, teammates, or competitive opponents. Furthermore, these existing techniques like finite state machines and behavior trees face limitations in handling the dynamically changing and unpredictable environments found in modern games, which can lead to repetitive behaviors or game bugs. To address these challenges and meet the increasing player demands for adaptive AI, researchers and game developers have begun to explore other technical options such as reinforcement learning (RL) [6].

RL is an interdisciplinary area of machine learning where an agent learns optimal actions by interacting with an environment to maximize cumulative rewards, guided by a policy and value function. It can be relatively straightforward to implement in new domains given a defined environment, set of actions, and reward structure [7]. Video games naturally provide these elements, making them ideal for RL applications. For instance, RL has been used for intelligent unit micro-management in *StarCraft* [8] and dynamical difficulty adjustment to maintain game flow in turn-based battle video games [9]. While RL is less suitable for large-scale strategic decision-making due to the space complexity, delayed rewards and other challenges, its effectiveness in smaller tactical scenarios shows promise for enhancing game AI [10].

When game environments become more complex, the knowledge space becomes too large for RL to store all state-action pairs. To solve this, researchers started to use neural networks to approximate state spaces or value/policy functions, a technique known as Deep Reinforcement Learning (DRL). A seminal work introduced the Deep Q-Network (DQN), which the agent learns directly from raw pixel data and achieves human-level performance in playing Atari games [11]. Later, DRL was applied to play FPS games and achieved human-level control [12], demonstrating its capability for real-time decision-making and navigation in 3D partially observable environments. Furthermore, DRL has also been used for navigation and pathfinding problems in AAA games [13], which outperformed traditional algorithms like A* [14]. Moreover, MuZero [15], an algorithm that integrates model-based planning with model-free learning, has demonstrated superhuman performance across various games, including Atari, Go, Chess, and Shogi. More recent applications of DRL include the study on *Gran Turismo Sport* (GTS) [16], a racing simulation game with realistic physics and competitive driving mechanics. It applied a soft actor-critic (SAC) algorithm to achieve superhuman performance in high-speed autonomous driving tasks [17]. Additionally, in 2023, researchers and developers from Electronic Arts integrated DRL into their automated testing pipeline for AAA titles such as *Battlefield*

*2042* and *Dead Space* [18].

While DRL has demonstrated significant success in video games, existing applications primarily focus on single-agent scenarios. In contrast, most modern video games involve multiple agents interacting cooperatively or competitively. Consequently, researchers and developers have turned to utilize MARL to develop AI systems that can effectively manage and optimize the interactions between multiple agents in games. In recent years, MARL has made significant progress. Early work on TD-Gammon in 1995 [19], demonstrated the potential of RL in a two-agent scenario. In this work, the AI agent learned to play Backgammon through self-play technique. Later on, the development of AlphaGo [20] in 2016 is another major milestone. AlphaGo applied supervised learning from human expert games and RL from self-play to master the game of Go, and eventually outperformed the world champion. Subsequently, AlphaGo Zero [21] in 2017 learned solely through self-play using RL without supervised learning. In the same year, AlphaZero [22] went to public, applied what it learned from Go to other board games such as Chess and Shogi. This demonstrated that a single algorithm (or model) could achieve superhuman performance across multiple games purely through self-play. Further breakthroughs were achieved with AlphaStar [23] in 2019, a significant advancement in MARL by mastering the well-known RTS game *StarCraft II*. AlphaStar trained using a combination of supervised learning from human games, DRL through self-play, and a novel League Training mythology, achieving a grandmaster level in *StarCraft II*. In the same time, OpenAI Five [24] showed their work of applying MARL to one of the most played MOBA game *Dota 2*. Unlike earlier work with two agents, OpenAI Five successfully trained multiple AI agents to play in a 5 vs. 5 setting and went on to defeat the world champion team OG at The International 2018. Given the increasing number of multi-agent systems in modern games and the progress in MARL along with its recent applications, focused reviews on this topic have become increasingly important. Existing reviews provide extensive insights into the theories, algorithms, and challenges of MARL [25]–[28]. Additionally, recent advances in deep learning across various video game genres have been reviewed [29], with specific discussions on DRL in video games [30], and the prospects of RL in the gaming industry [31]. While these surveys primarily emphasize single-agent scenarios, we aim to provide a comprehensive review of MARL in video games, highlighting challenges, recent advancements, applications, and future directions.

This paper aims to provide a comprehensive review of MARL applications and research in video games. We will discuss implementation challenges such as partially observable and nonstationary environments, delayed and sparse rewards, team incentive mechanisms, communication and coordination, credit assignment and scalability. Furthermore, we propose a novel method to estimate game complexity using five key dimensions: Observability, State Space, Action Space, Reward Sparsity, and Multi-Agent Scale. We will examine notable studies and successful implementations, starting with two-agent games such as Backgammon, Go and *Blade & Soul*. We will then extend our review to multi-agent games involving more than two agents from simple to more complex games categorized by their genres, which includes the most popular one: Competitive and Sports games such as *3v3 Snake*, *Google Research Football*, *Ubisoft's Roller Champions*, and *Rocket League;* FPS games including *Doom* and ViZDoom, *Minecraft*, and *Quake III Arena: Capture the Flag;* RTS and MOBA, with games like *StarCraft II*, *Dota 2*, and *Honor of Kings*. We will discuss challenges from both the game development and technical perspectives and propose new directions for future research. Through this review, we hope to inspire further research and innovation in game AI, and ultimately advancing the capabilities of both MARL and game AI systems.

## II. BACKGROUND

### A. Terminology

To establish a consistent conceptual foundation for this review, we define key terms related to MARL in the context of video games.

- An **agent** is defined as "a computer system that is situated in some environment and capable of autonomous action in order to meet its design objectives" [34]. Operationally, within RL, an agent is characterized explicitly by its *observations* (sensory inputs), *state* (information about the environment), *actions* (outputs via actuators), and *rewards* (feedback) [14], [34].
- A **multi-agent system** is a system that consists of a number of distributed agents, which communicate and interact with one another, typically by exchanging messages through some computer or network infrastructure [34].
- **Non-player characters (NPCs)** are autonomous character entities in video games that are not controlled by the player. Traditional NPCs often rely on scripted, rule-based systems such as finite state machines or behavior trees [35].

This review focuses explicitly on learning-based NPCs implemented as RL agents, rather than depending on predefined rules or logic scripts.

### B. Reinforcement Learning (RL)

In typical or single-agent RL, the environment is stated in the form of a **Markov Decision Process** (MDP), which is a mathematical model defined by a tuple of five elements $(S, A, P, R, \gamma)$. At each time-step, the agent observes the current state $s_t$, selects an action $a_t$, receives a reward $r_{t+1}$, and transitions to the next state $s_{t+1}$ based on the transition probability $P(s_{t+1}|s_t, a_t)$. The agent's goal is to learn a policy $\pi(a_t|s_t)$, which maps states to actions, in order to maximize the cumulative reward, or return, $R_t$, defined as:

$$R_t = \sum_{i=0}^{\infty} \gamma^i r_{t+i} \qquad (1)$$

where $\gamma$ is the discount factor in [0, 1] that determines how much future rewards are weighted compared to immediate rewards [6].

One of the fundamental algorithms in RL is **Q-Learning** [32], which an agent learns the optimal policy by estimating the value of state-action pairs, represented by a function $Q(s, a)$.

The goal is to learn the Q-function, which predicts the total expected reward for taking action $a$ in state $s$ and following the optimal policy thereafter. The Q-values are updated iteratively using the Bellman equation:

$$Q(s_t, a_t) \leftarrow Q(s_t, a_t) + \alpha \left[ r_{t+1} + \gamma \cdot \max_a Q(s_{t+1}, a) - Q(s_t, a_t) \right] \quad (2)$$

where $\alpha$ is the learning rate. On the other hand, policy gradient [33] methods take a policy-based approach by directly learning the policy $\pi_\theta(a|s)$ without needing to compute value function. The policy is parameterized by $\theta$, and the objective is to maximize the expected cumulative reward $J(\theta)$:

$$J(\theta) = \mathbb{E}_{\pi_\theta}\left[\sum_{t=0}^{\infty} \gamma^t r_t\right] \qquad (3)$$

where $\gamma$ is the discount factor that controls the importance of future rewards.

### C. Deep Reinforcement Learning (DRL)

DRL is a combination of RL and deep learning, where neural networks are used to approximate policies or value functions. One widely-used architecture is the **Convolutional Neural Network** (CNN) [11], where the agent learns from visual inputs such as images or pixel data. Moreover, in environments with sequential data, **Recurrent Neural Networks** (RNNs) and their variant, **Long Short-Term Memory** (LSTM) [36], [37] networks, are often used.

Building on Q-Learning in RL, **Deep Q-Leaning** or DQN [11] extends the algorithm by using a deep neural network to approximate the Q-function, so it can scale to high-dimensional state spaces. Instead of maintaining a Q-table for each state-action pair, DQN typically uses a CNN to estimate the Q-values for different actions based on raw pixel input. To stabilize learning, DQN incorporates experience replay, which stores and reuses past experiences to break correlations between consecutive samples, and target networks, which help reduce oscillations by keeping a separate, slowly-updating network for generating target values [38].

Moreover, the **Actor-Critic** [39] architecture extends policy gradient methods by combining the benefits of policy-based and value-based approaches**.** In this framework, the actor learns the policy $\pi_\theta(a|s)$, mapping states to actions, while the critic estimates the value function $V(s)$, providing feedback to the actor to improve the policy. This approach reduces the high variance typically present in policy gradient methods by leveraging value-based learning to guide policy updates. The purpose of the actor is to maximize the expected cumulative reward, but instead of using raw returns, it utilizes feedback from the critic through the advantage function**:**

$$\nabla_\theta J(\theta) = \mathbb{E}_t[\nabla_\theta \log \pi_\theta(a_t|s_t) \cdot A(s_t, a_t)] \qquad (4)$$

where $\pi_\theta(a_t|s_t)$ is the policy parameterized by $\theta$, mapping states to actions with the advantage function defined as:

$$A(s_t, a_t) = Q(s_t, a_t) - V(s_t) \qquad (5)$$

which measures how much better the action $a_t$ is compared to value function $V(s)$ and minimizes the temporal difference error to improve its estimation over time:

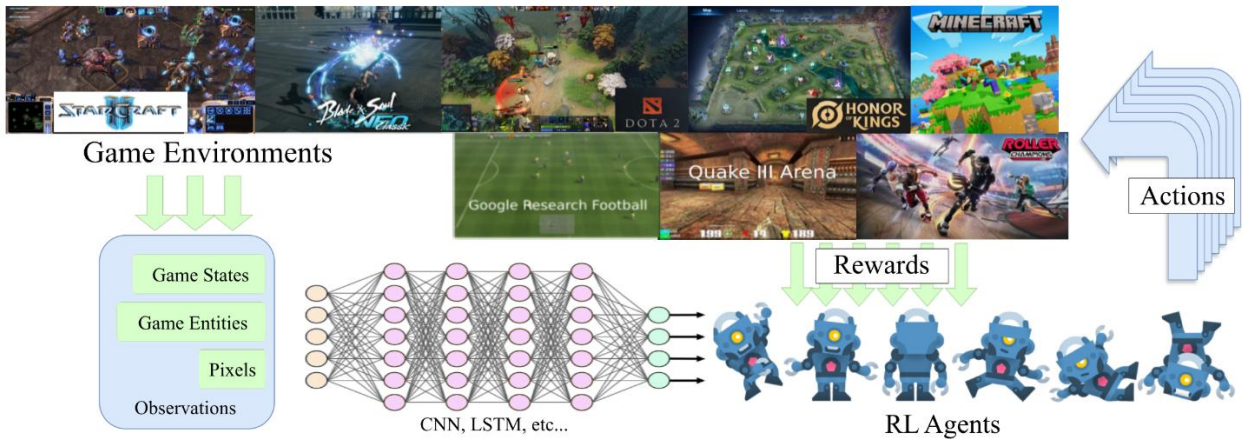

**Fig. 2**. This diagram shows the typical flow of MARL in video games. Agents receive observations, process them through different neural networks, and output actions that lead to the next state in the game environment.

$$\delta_t = r_{t+1} + \gamma V(s_t + 1) - V(s_t). \tag{6}$$

**Deep Deterministic Policy Gradient** (DDPG) [40] extends the actor-critic framework to continuous action spaces using a deterministic policy:

$$a = \mu_\theta(s) \tag{7}$$

Unlike stochastic policy gradient methods, DDPG deterministically selects the best action for a given state. It uses off-policy learning to maintain sample efficiency, where the agent stores experiences in a replay buffer and learns from them. The critic minimizes the Bellman error to stabilize value estimation:

$$y = r + \gamma Q'[s', \mu'(s')] \tag{8}$$

**Proximal Policy Optimization** (PPO) [41] improves on traditional policy gradient methods by ensuring stable and reliable policy updates. It introduces a clipped surrogate objective to prevent large policy updates during each iteration, balancing exploration and exploitation. This method is widely used in both discrete and continuous action spaces due to its stability and efficiency, with its clipped surrogate objective function defined as:

$$L_{CLIP}(\theta) = E_t [min(r_t(\theta)A_t, clip(r_t(\theta), 1 - \epsilon, 1 + \epsilon)A_t)] \tag{9}$$

and the final PPO objective function as:

$$L_{PPO} = L_{CLIP} - c_1 L_{value} + c_2 L_{entropy} \tag{10}$$

### D. Multi-Agent Reinforcement Learning (MARL)

MARL builds on RL and DRL by extending the framework of MDP to environments involving multiple agents [42]. In MARL, agents interact with both the environment and each other, each following its own policy, as shown in **Fig. 2**. Formally, MARL is modeled as a **Markov Game** (also known as a Stochastic Game) [43], defined by the tuple:

$$T = (S, \{A_i\}_{i=1}^N, P, \{R_i\}_{i=1}^N, \gamma) \tag{11}$$

where $S$ is the state space shared by all agents. $A_i$ is the action space of agent $i$, $P(s'|s, a_1, \ldots, a_N)$ is the transition probability function, which describes how the state evolves given the joint actions of all agents. $R(s, a_1, \ldots, a_N)$ is the reward function for agent $i$ which may depend on the joint actions and the state. $\gamma$ is the discount factor that balances immediate and future rewards.

The goal of each agent is to learn a policy $\pi_\theta(a|s)$ that maximizes its cumulative discounted reward:

$$J_i(\pi_i) = \mathbb{E} \left[ \sum_{t=0}^\infty \gamma^t R_i(s_t, a_t) | \pi_1, \ldots, \pi_N \right] \tag{12}$$

Depending on the task or goal they address, there are three types of multi-agent interactions [43], [44]:

1) **Competitive**
   These interactions are typically modeled as Zero-Sum Markov Games, where the sum of all agents' rewards is zero, implying that one agent's gain is another's loss.

2) **Cooperative**
   In cooperative settings, agents share a common reward function $R(s, a_1, \ldots, a_N)$, aligning their objectives. This setup is referred to as a multi-agent MDP and emphasizes coordination for team-wise optimal outcomes.

3) **Mixed**
   It is also referred as general-sum games. These settings combine competitive and cooperative elements, where agents may cooperate temporarily while pursuing their individual objectives.

**Self-play** is a widely adopted training method in MARL, particularly in competitive settings where agents learn by playing against various versions of themselves. The method allows agents to iteratively improve their performance without the need for external opponents or pre-collected data. This approach was notably successful in TD-Gammon [19]. Later, AlphaGo [20] and AlphaZero [21] demonstrated the power of self-play combined with DRL and Monte Carlo Tree Search to master complex board games like Go, chess, and shogi. More recently, self-play has been widely applied in various video games such as *StarCraft II* [23] and *Dota 2* [24] as a technique to improve agent performance and training efficiency.

Another fundamental aspect of MARL is the choice of training and execution paradigm. The three main paradigms are:

1) **Centralized Training, Centralized Execution (CTCE):**
   All agents are trained and executed using shared global information. This setup allows tight coordination but is rarely practical for real-time systems due to scalability and communication constraints.

2) **Centralized Training, Decentralized Execution (CTDE):** Agents are trained with access to global state or other agents' information, but at execution time, they operate using only local observations. This paradigm maintains a balance between coordination and deployability and is widely used in practice.

3) **Decentralized Training, Decentralized Execution (DTDE):** Each agent is trained and executed using only its own observations and experiences. This setup is highly scalable and realistic but can make coordination significantly more difficult.

Across the reviewed games, the CTDE paradigm dominates as the preferred training methodology. This reflects a common design requirement that agents are trained with access to centralized or shared state information to enable coordinated policy learning. Also, decentralized execution complies with real-time performance constraints. CTDE is particularly effective in environments where agents need to cooperate while still operating independently during gameplay. A few exceptions adopt DTDE, such as ViZDoom's Deathmatch, which benefits from fully independent training due to its simple, fully competitive setup. In highly flexible environments like *Minecraft*, both CTDE and DTDE appear depending on the scenario and mini game. In contrast, CTCE is typically impractical for real-time video games, where latency, scalability, and individual agent autonomy are essential.

## III. METHODOLOGY

This section outlines the methodology used to collect, screen, and include the literature and work analyzed in this review. Our objective was to construct a representative and relevant corpus of works on MARL in video games. The final corpus comprises 84 reports, separated into 40 core studies used for synthesis and analysis, and 44 supplementary works included to support background, theoretical framing, and methodological context.

### A. Scope and Structure

This review was designed to systematically identify, organize, and analyze research on the application of MARL in video game environments. The scope of our review is limited to works in which MARL techniques are implemented and evaluated in either competitive or cooperative multi-agent video game settings. We include both two-agent and multi-agent environments but exclude works that focus solely on single-agent systems or other non-video-game environments. The core synthesis is based on a curated set of 40 primary studies, identified through expert knowledge, professional search databases and citation chaining. To support technical framing and historical context, an additional 44 documents are included for background.

### B. Data Sources and Search Strategy

The identification of relevant literature was guided first by professional experience and domain familiarity accumulated through ongoing research and development in MARL.

We then performed a structured search across major academic databases, including IEEE Xplore, ACM Digital Library, SpringerLink, Elsevier ScienceDirect, Wiley Online Library, and arXiv, to collect high-quality studies. These platforms were selected for their relevance to both AI and game technology domains. Search terms included combinations of "multi-agent", "reinforcement learning", and "video games", tailored to each platform's query syntax. This search phase emphasized peer-reviewed journal articles, conference proceedings, and preprints from reputable institutions.

To ensure comprehensive coverage beyond canonical databases, we also conducted a supplementary search using Google Scholar, which indexes broader sources such as industry research and non-indexed preprints. The query used was: "multi-agent", "reinforcement learning", and "video game". Google Scholar returned over 17,000 results, with only the first 1,000 accessible. From these, the top 500 entries (25 pages) sorted by relevance were manually screened.

Finally, we applied forward and backward citation chaining (snowballing) on key papers identified during earlier phases. This step allowed us to capture important studies that may have used non-standard terminology or were not ranked highly by search algorithms.

### C. Inclusion and Exclusion Criteria

To ensure consistency and relevance in the reviewed corpus, we applied tailored inclusion and exclusion criteria depending on the intended role of each report within the structure of the review.

1) **Inclusion Criteria** - Reports were included in the core corpus if they met all of the following criteria:
   - They present a method involving MARL and applied MARL to one or more video game environments, including both real-time and turn-based games, commercially or academically.
   - They provided implementation details, such as model architecture, training algorithm, reward structure, or experimental results.
   - These studies align with the definitions established in Section II.A.

2) **Exclusion Criteria** - Reports were excluded if they met any of the following criteria:
   - Focused solely on single-agent RL or no RL.
   - Employed abstract simulations or robotics tasks not involving a recognizable game environment.
   - Were tutorial, visionary, or purely theoretical in nature without implementation or evaluation.

3) **Additional Source Filtering** – All Reports retrieved were subject to the same criteria. For preprints, inclusion was limited to reports that:
   - Reported substantial original empirical results and provided full methodological transparency.
   - Were affiliated with reputable research institutions or accepted by peer-reviewed workshops.

4) **Background and Supporting Documents** - A separate set of reports was manually selected to provide foundational context and theoretical grounding. These include works introducing core RL algorithms, general

MARL theory, and relevant surveys or industry references. These reports were not selected through the systematic search process and are not included in the comparative synthesis, but are cited as needed.

### D. Discussion of Strengths, Weaknesses, and Biases

This review uses a multi-stage and reproducible methodology based on professional expertise, structured database searches, and clear inclusion criteria. However, citation chaining can bias selection toward more visible or recent studies. Manual filtering also introduces subjectivity. We mitigate these limitations by transparently documenting sources and selection criteria.

## IV. MARL CHALLENGES IN GAMES

### A. Nonstationary and Partially Observable Environments

Modern game environments are inherently nonstationary due to the presence of multiple agents interacting within the same environment. This violates the stationary assumption of traditional MDPs, where transition probability and reward functions are expected to remain stationary. Moreover, video game environments often only partially observable because they typically have large maps that cannot be fully captured within a single game camera view. In addition, many games also have mechanics like "fog of war" to add strategic depth, where areas and opposing units on the map remain unobservable unless nearby ally units provide vision [45]. This partial observability significantly complicates the learning process, requiring Partially Observable MDPs (POMDPs) [37], [46]. Consequently, many popular algorithms, such as independent DQN, become impractical without substantial modifications. Because of these elements, designing such MARL system in *StarCraft II* is extremely challenging, due to uncertainty and incomplete information [26].

### B. Delayed and Sparse Rewards

In video games, the issue of delayed and sparse rewards is significant due to the real-time nature and length of gameplay. In general, video games run at 16-60 frames per second, and one game could last from minutes to hours depending on the game. For example, a game of *Dota 2* lasts approximately 45 minutes and runs at 30 frames per second [24]. The OpenAI Five agent selects an action every fourth frame, summing up to roughly 20,000 steps per episode, compared to only 150 steps in a game of Go. Additionally, in *StarCraft II*, professional players can perform as many as 500 actions per minute (APM) [45], with matches usually lasting around 15 minutes. These long game durations plus high action frequencies can cause agents to receive infrequent feedback from their actions.

### C. Designing Team Incentive Mechanisms

In multiplayer video games, competitive settings typically involve team-wise zero-sum Markov games, such as destroying opponents' bases in *Dota 2*, *StarCraft II*, or *Honor of Kings*. Cooperative settings can include solving team puzzles, as in *Minecraft*, or collaborative play in online games like *World of Warcraft* [47]. In both scenarios, teamwork weighs much more than individual abilities in order to achieve objectives. To encourage teamwork, challenges arise when designing incentive mechanisms that align individual agent incentives with overall team objectives. These mechanisms must prevent agents from pursuing actions that benefit themselves at the expense of the team's success, but instead encourage them to work together for the greater success.

### D. Communication and Coordination

Furthermore, communications play a vital role in coordination among agents, and they are critical for achieving optimal policies. In many video games, agents operate in partially observable environments, and effective communication allows them to share their local observations, forming a more comprehensive understanding of the environment. Additionally, agents often need to make joint actions, coordinating their efforts to execute macro-level strategies such as coordinated attacks, defense, or resource management [48]. Also, effective communication is essential for aligning individual policies and making micro-level decisions. This involves sharing local observations, intentions, and plans in real-time.

### E. Credit Assignment

The multi-agent credit assignment problem [49], [50] is another significant challenge in MARL for video games. This issue is especially evident in strategy games with long time dependencies. It is challenging to track when macro-level strategies, such as terrain control or expansion, are executed over several minutes with hundreds of steps each minute. Additionally, when such a macro-level strategy is successfully completed by all agents as a team, it becomes hard to determine the contribution of each agent to the team's success. More specifically, designing a reward structure that appropriately credits the series of micro-level actions taken by individual agents remains a complex task.

### F. Scalability and Computational Efficiency

As the number of agents in a game increases linearly, the complexity of managing their interactions increases exponentially. Scalable algorithms are needed to handle large-scale multi-agent environments. For instance, while state space for Go is estimated at $10^{170}$, MOBA can reach as high as $10^{20,000}$ [48]. Moreover, AlphaStar accumulated 200 years of playing *StarCraft II* [23], and in *Dota 2*, OpenAI Five produced ~180 years of gameplay data per day with 128,000 CPU cores and 256 GPUs. Even with the use of such extensive CPU and GPU resources, it only supported up to 17 heroes out of 117, equivalent to 14.57% of total capabilities [24]. In addition, the real-time nature makes it more computationally demanding and challenging. As discussed earlier, in *StarCraft II*, professional players make up to 500 APM [45], meaning a decision is made every 125 milliseconds. To achieve the performance, a MARL agent must make an optimal decision close to that time.

## V. MARL-BASED COMPLEXITY FRAMEWORK

### A. Limitations of Traditional Approaches

Traditional approaches to classifying games typically rely on subjective or production-oriented measures, such as development budget, team size, or informal industry-standard categories (AAA, AA, Indie). However, these approaches often correlate poorly with the inherent complexity of the game from a learning perspective. Budget and team size may indicate production value or content volume, but they do not necessarily reflect the strategic and cognitive difficulty presented by the game environment. Additionally, relying on subjective evaluations of complexity by human players is problematic, as individual player experience, skill level, and familiarity with game mechanics introduce variability that undermines reproducibility and objective measurement.

In contrast, RL theory provides a structured and consistent measure for evaluating learning difficulty. While human cognition varies widely, RL environments are typically formalized using MDP. This approach enables consistent reasoning about how complex a game is to learn, play, and master.

### B. Dimensions of the Proposed Framework

To systematically assess complexity within this learning-oriented context, we propose a classification based on five fundamental dimensions derived from the MDP framework:

1) **Observability**: The degree to which agents can perceive the full state of the environment at each decision point.
2) **State Space**: The size and format of the observation available to the agent at each decision point, used as a proxy for environment complexity.
3) **Action Space**: The number and type of actions an agent can choose from at each decision point.
4) **Reward Sparsity**: The frequency and distribution of learning feedback received by agents in response to meaningful actions.
5) **Multi-Agent Scale:** The number of interacting agents in multi-agent environments.

These dimensions capture core aspects of agent-environment interaction and form an effective framework for evaluating game complexity from a learning perspective.

### C. Scope and Limitations

Real-time video games represent some of the most challenging environments for agents to learn, as they are typically *partially observable*, *multi-agent*, *stochastic*, *sequential*, *dynamic*, *continuous*, and *unknown* [14]. Developing a comprehensive, universally applicable measure of game complexity is therefore beyond the scope of this review. Instead, our objective is to introduce a practical and reproducible classification that facilitates meaningful comparisons and deeper understanding of MARL performance across diverse game environments.

In the following sections, we apply this MARL-based game complexity framework to systematically review and analyze MARL applications across various game genres, structured in order of increasing complexity.

## VI. MARL IN TWO AGENT GAMES

As the simplest form of multi-agent interactions, two-agent games also fall into the categories of cooperative, competitive, and mixed settings [43], [44]. Among these, competitive two-agent games have the most attention, particularly when they outperform human players. We will begin our review with two-agent games, transitioning from foundational works like Backgammon and Go to real-time video games. While Backgammon and Go are neither video games nor real-time environments, they have significantly contributed to advancing MARL research.

### A. Backgammon & TD-Gammon

Backgammon is a strategic two-player game where the objective is to move all one's checkers off the board based on dice rolls. In 1990s, a significant breakthrough in the RL domain was achieved with the application of temporal-difference (TD) learning to master the game [51]. Using a multi-layer perceptron (MLP) neural network, TD-Gammon approximated the value function V(s) to predict the probability of winning from any given game state and achieved master-level play that is extremely close to the world's best human players.

The state space in Backgammon, roughly estimated at $10^{20}$ possible configurations [51], and the action space, with approximately $10^4$ possible moves per turn. TD-Gammon employed a MLP neural network, which consisted of an input layer that processed the board configuration, followed by hidden layers that captured the patterns of optimal play, and an output layer that estimated the value of the position. Additionally, its training process used self-play technique. This allowed the system to refine strategies and policies autonomously without human involvement. It significantly improved training efficiency by enabling TD-Gammon to generate its own data, accelerating learning by exploring diverse game states and refining its strategies over time. Central to this learning process was the TD($\lambda$) algorithm, which updated the network's value estimates based on the temporal difference error function:

$$\delta_t = R_{t+1} + \gamma V(s_{t+1}) - V(s_t) \tag{13}$$

where $R_{t+1}$ is the reward at time t+1 and $\gamma$ is the discount factor.

The impressive results of TD-Gammon demonstrated a level of play competitive with top human experts, occasionally surpassing world-class players. This pioneering integration of RL and neural networks showcased the potential of these techniques to handle stochastic environments. Moreover, the success of self-play in TD-Gammon established a foundation for MARL training processes, influencing seminal projects such as DQN in Atari 2600 games [11] and AlphaGo [20].

### B. Go & AlphaGo Series

Go is an ancient two-player competitive strategy game that originated in China over 2,500 years ago. In the game, players on two sides take turns placing black or white stones on a 19x19 grid to control the largest area on the board and capture the opponent's stones. In 2016, DeepMind developed AlphaGo [20] marked a significant milestone in applying DRL to the game. AlphaGo combined supervised learning with RL through self-play to learn to play the game and used Monte Carlo Tree Search (MCTS) to explore potential future moves. As a result, it achieved a victory against the European Go champion by 5 games to 0.

TABLE I: MARL OVERVIEW IN TWO-AGENT GAMES

| Game | Real-Time vs Turn-Based | Observability | Learning Approaches | Deterministic vs Stochastic |
|---|---|---|---|---|
| **Backgammon & TD-Gammon** | Turn-Based | Fully | TD($\lambda$), Self-play | Stochastic |
| **Go & AlphaGo** | | | MCTS, CNN | Deterministic Game, Stochastic AI |
| **B&S** | Real-Time | Partially | PPO, LSTM | Stochastic |

In the previous section, we reviewed Backgammon, which has a state space of $10^{20}$. In comparison, the state space of Go is much larger, with an estimated $10^{170}$ possible board configurations, making it one of the most challenging board games. The action space is also large, with hundreds of available moves at any given turn. AlphaGo addressed these challenges using a CNN to process the board state and a policy network to prioritize promising moves. The training process involved two phases: supervised learning on a dataset of 30 million positions from the KGS Go Server with an accuracy of 57.0%, and DRL through self-play. The MCTS algorithm played a crucial role in this process by simulating numerous future move sequences, guiding the policy network by updating the value of each node based on simulated game outcomes.

AlphaGo Zero and AlphaZero further advanced this approach by eliminating the need for human data and relying solely on DRL from self-play. These systems used a single neural network to evaluate positions and select moves, learning from scratch with the same MCTS framework. The loss function combined policy loss and value loss to guide training:

$$L(\theta) = (z - v)^2 - \pi^T \log P + c\,||\theta||^2 \qquad (14)$$

where $z$ is the game outcome, $v$ is the value prediction, $\pi$ is the MCTS-based policy, $P$ is the policy prediction, and $c$ is a regularization parameter. This approach achieved significant improvements, with AlphaGo Zero surpassing the original AlphaGo, and AlphaZero generalizing the approach to other games like Chess and Shogi.

AlphaGo's success paved the way for advancements in MARL, demonstrating the effectiveness of self-play and the use of CNNs within DRL to process large observation space. Techniques from AlphaGo were widely adopted by other games, such as *StarCraft II* (AlphaStar), *Dota 2* (OpenAI Five) and *Honor of Kings* and beyond.

### C. *Blade & Soul*

AI research has traditionally focused on turn-based games like Backgammon and Go, where agents have unlimited time to compute optimal strategies using algorithms like MCTS. In contrast, real-time games like *Blade & Soul* require agents to make quick decisions continuously within milliseconds and often with imperfect information.

As one of the real-time games, *Blade & Soul* (B&S), developed by NCSOFT, is known for its fun action combat mechanics and multiplayer environment. The game also has a one-on-one fighting mode where two players each controls a character and fight against each other in a given time (**Fig. 3**). To explore the potential of AI in the game, the developers integrated DRL and achieved pro-level AI agents capable of competing against human players in this mode [52].

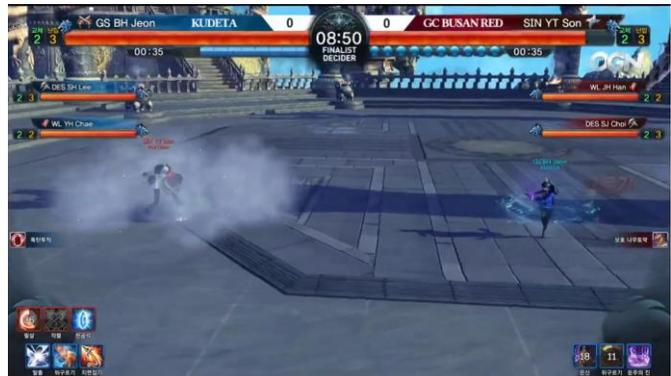

Fig. 3. *Blade & Soul* 1 vs. 1 Tournament.

Unlike traditional fighting games in 2D format, B&S has a vast action space with high dependencies between moves, skills, and strategies in 3D. It focuses exclusively on competitive multi-agent interactions. The neural network architecture used for training consists of an LSTM-based model with four heads, each responsible for different aspects of decision-making: skill selection, movement, targeting, and evaluation. The game's state space includes detailed observations such as hit points (HP), skill cooldowns, and opponent positions, while the action space covers 44 skills, movement, and targeting directions. The training process utilizes an actor-critic off-policy algorithm, which handles policy lag through truncated importance sampling. A self-play curriculum with diverse opponent pools was developed to ensure the agent could adapt to various competitive strategies. Additionally, the training incorporated "data-skipping" techniques to improve data efficiency by discarding passive "no-op" actions, enhancing the AI agents' ability to explore and optimize its decision-making.

The AI agents trained using this method demonstrated pro-level performance, achieving a 62% win rate against professional gamers in the 2018 B&S World Championship. The aggressive agent, in particular, outperformed all human opponents in both live events and pre-tests. These results demonstrate the potential of DRL in mastering complex real-time competitive games. The methodologies developed for B&S can be generalized to other two-player competitive games, providing valuable insights for future AI research and game development [52].

The exploration of RL in two-agent games has revealed the depth and diversity of this domain, spanning from early successes in Backgammon with TD-Gammon to the groundbreaking achievements of the AlphaGo series in Go, and extending to real-time video game with *Blade & Soul*. To sum up, TABLE I provides an overview of the key characteristics

TABLE II: Overview of the complexity of various video games.

| Genre | Game | Observability | State Space | Action Space | Reward Sparsity |
|---|---|---|---|---|---|
| **Competitive and Sports Games** | *3v3 Snake* | Full | 12×10×20 matrix | 4 options per step, 200 steps | Frequent |
| | *Google Research Football* | High | 115 floats | 19 options per step, 3000 steps | |
| | *Roller Champions* | | 78 game entities | 9 options per step, ~5400–12600 steps, | |
| | *Rocket League* | | 3 arrays of game states (Lucy-SKG) | 90 options per step, 9000 steps | |
| **FPP and FPS Games** | *Doom & VizDoom* | Partial | 320x240 RGB pixels | 8 options per step, ~21,000 steps, | Intermediate |
| | *Minecraft* | | Varies by platform | Varies by platform | |
| | *Quake III Arena Capture the Flag* | | 84x84 RGB pixels | 540 options per step, 4,500 steps, | |
| **RTS and MOBA Games** | *StarCraft II* | Partial | 512 units with 14 attributes, 128x128 grid map 3 attributes of player data 32x20 camera | $10^{26}$ options per step, ~14,400 steps | Sparse |
| | *Dota 2* | | ~16,000 per observation | 8000–80000 options per step, ~ 81,000 steps | |
| | *Honor of Kings* | | 9,227 scalar features, 6×17×17 spatial features | 10 options per step, ~20,000 steps | |

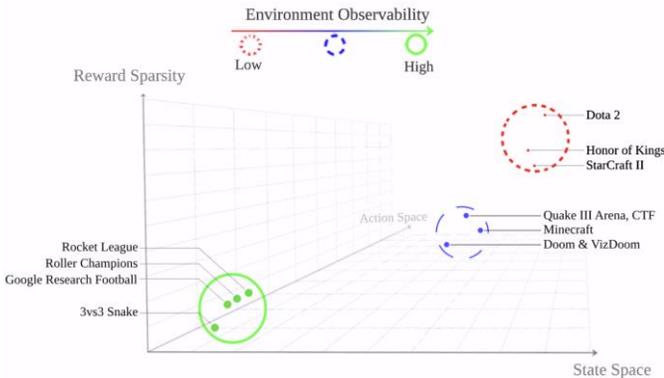

**Fig. 4.** Categorization of video games based on their complexity, measured by state space, action space, reward sparsity, and environment observability. The figure illustrates games grouped into clusters reflecting their complexity.

and learning approaches used in the two-agent games discussed.

While competitive settings have drawn significant attention due to their superhuman performance, it is also important to recognize that two-agent games are not limited to competitions. Cooperative scenarios, such as those studied in the game *Overcooked*, highlight the potential for evaluating and enhancing human-AI cooperation [53]. This area of research is advancing the development of cooperative agents in game AI, such as companion NPCs, and includes studies focused on cooperative play between AI and human players.

## VII. MARL IN MULTI-AGENT GAMES

Building on the discussion of two-agent games, we now turn our attention to multi-agent environments involving more than two agents. Unlike two-agent scenarios, these games often require coordination within teams. As detailed in Section V, we use a MARL-based framework to estimate game complexity ordered by increasing learning complexity. A summary of this classification is provided in **Fig. 4** and TABLE II.

### A. Competitive and Sports Games

Competitive and sports games are characterized by their structured, rule-based gameplay and emphasis on immediate objectives, such as scoring points or winning matches. Unlike genres that require long-term strategic planning or navigation through complex environments, competitive and sports games focus on precision, timing, and execution within more confined settings. Players in these games operate under clear, well-defined rules, engaging in fast-paced actions that require quick decision-making and coordination. The environments are typically simpler and more predictable than other genres with combat gameplay, allowing for a direct, skill-driven experience where success is determined by the player's ability to adapt rapidly to the game's immediate demands.

#### 1) *3v3 Snake*

As one of the well-known real-time games, *3v3 Snake* extends the classic *Snake* game into a multi-agent environment. In its general mode, multiple teams of snakes compete to grow the longest by consuming beans while avoiding collisions with themselves, their teammates, or their opponents. In one study [54], the game is configured with two teams of three snakes each on a 10x20 fully observable map with toroidal boundaries, where a snake that crosses one edge of the map reappears on the opposite edge, with this wrapping behavior applied to both horizontal and vertical boundaries. *3v3 Snake* is particularly suitable for MARL research, presenting intriguing puzzles related to both teamwork and competition with a simple fully observable environment.

The AI development for *3v3 Snake* uses a rule-enhanced MARL algorithm that integrates traditional rule-based strategies with advanced RL techniques [54]. Its neural network architecture is built using eight residual blocks, each containing two 3x3 convolutional layers for efficient feature extraction and decision-making. This architecture supports both the policy network, which guides the snakes' actions, and the value network, which evaluates the game state, sharing a common structure. The observation space is represented by a 12-channel matrix, capturing key game elements such as snake positions, bean locations, and a territory matrix that incorporates human-derived rules. The action space is limited to 4 directional movements with the opposite direction always being illegal. Its training process employs Distributed PPO across 20 actor processes and one learner. A novel aspect of this approach is the integration of the territory matrix within the MARL framework, allowing the AI agents to apply human-like strategic insights.

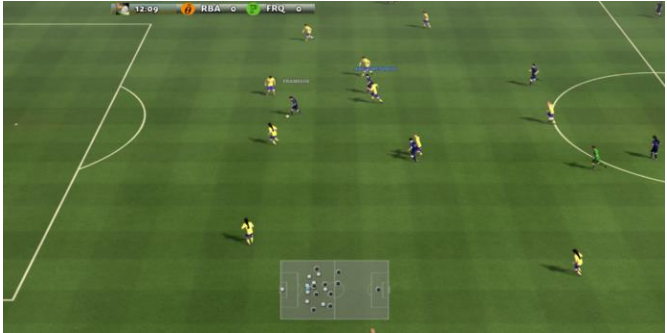

**Fig. 5**. *Google Research Football* Player's View.

In the context of MARL, *3v3 Snake* provides a collaborative environment so they must learn to cooperate within their team while simultaneously competing and defending the opposing team. The rule-enhanced MARL algorithm developed for this game demonstrates the importance of integrating human-designed rules with learned policies to achieve superior performance. The trained agents consistently outperform both rule-based algorithms and human players, highlights the potential of *3v3 Snake* for future MARL research.

### 2) *Google Research Football*

*Google Research Football* (GRF) is a physics-simulated 3D video game designed to replicate professional football (soccer) as shown in **Fig. 5**. It supports all major actions seen in a real football game and is highly customizable and flexible for RL research. The environment allows researchers or players to either control an entire team of players as a single agent or engage in MARL scenarios where multiple agents manage different players on the same team.

The platform is also optimized for computational efficiency, incorporates stochastic elements, and is integrated with commonly known RL models, while also supporting cooperative MARL elements such as agent communication. In this environment, opponents can be either built-in AI bots or other trained agents in multiplayer settings, providing a versatile and challenging context for the development and evaluation of RL and MARL algorithms.

The GRF environment offers an observation space composed of three distinct representations: Pixels, Super Mini Map (SMM), and Floats. The Pixel representation consists of a 1280×720 RGB image processed by a CNN. The SMM utilizes four 72×96 binary matrices to capture positional information about players, the ball, and the active player. Lastly, the Floats representation provides a compact 115-dimensional vector of key game metrics, such as player coordinates, ball possession and direction, active player, and game mode. This diversity in observation allows researchers to experiment with varying levels of abstraction in agent training. The action space is discretized, encompassing movement in 8 directions, passing, shooting, sprinting, sliding, and dribbling. GRF implements two reward functions: the SCORING function rewards agents based on successful goals, while CHECKPOINT addresses the reward sparsity of SCORING by offering intermediate rewards as the ball progresses toward the opponent's goal. For training, GRF integrates three state-of-the-art RL algorithms: IMPALA [55], PPO [41], and Ape-X DQN [56]. Additionally, GRF includes the "Football Academy", a suite of predefined scenarios that progressively increase in difficulty, allowing agents to systematically train and refine their skills.

In the context of MARL, GRF environment supports both cooperative and competitive multi-agent setups, ranging from small-scale scenarios to full 11 vs. 11 matches. It provides a versatile and computationally efficient platform for advancing RL and MARL with a rich and diverse observation space, well-defined action space, and hierarchical reward structures. Its integration of state-of-the-art algorithms and structured training scenarios makes it a valuable tool for developing and testing MARL in a realistic football simulation.

### 3) *Ubisoft's Roller Champions*

In addition to GRF, another significant platform is the Unity ML-Agents Toolkit [57]. A notable game of its application is *Roller Champions*, a third-person perspective, fast-paced, team-based competitive multiplayer sports game developed by Ubisoft. In the game, players skate around an oval-shaped arena with the objective of scoring goals by throwing a ball through a hoop. Similar to other sports games, it emphasizes skill-driven gameplay centered around one objective (the ball) with clearly defined rules. The game features both coordination and competition, as players must pass the ball, defend, and position themselves strategically to complete laps and score points.

To enhance player experience, the developers integrated MARL into *Roller Champions* AI systems with the goal of creating agents that can effectively compete against and collaborate with human players [58]. The system was not designed solely to achieve superhuman performance but to contribute to the overall enjoyment of the game by fostering cooperative strategies, maintaining game balance, adapting to different levels of player skill, and replacing players when they disconnect. This practical implementation of MARL features very efficient deployment and training processes. This allows the AI agents to quickly adapt gameplay and balance changes, supporting rapid agile game development. By focusing on fun and varied gameplay rather than just win rates, its AI systems contribute to a more immersive and rewarding experience for players across all game modes. Moreover, this practical approach to diversified MARL research sets valuable examples for other researchers and developers, demonstrating how MARL can be effectively integrated into modern game design to enhance both player engagement and development efficiency.

The AI system for *Roller Champions* utilizes PPO as its core learning algorithm, with a policy network consisting of three layers, each with 512 neurons. The decision interval is set at 15 Unity's FixedUpdates [59], corresponding to 0.02 seconds per update under default settings. Training is conducted through self-play across 3–15 simultaneous environments against a wide range of strategies and skill levels. The process is balanced between efficiency and performance, allowing for the rapid development of new models. Overall, it takes only 1–4 days to produce a new model following gameplay or balance changes, making this approach highly suitable and practical for fast-paced, agile game development in a live service game.

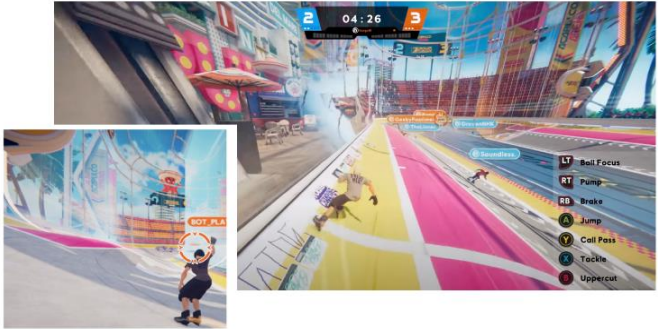

**Fig. 6.** A screenshot of *Roller Champions* in 3v3 mode, showing a score of 2-3 with remaining time and player controls (available actions). The mini view shows the Orange agent strategically positioned at the goal, performing a Call for Pass as their teammate passes checkpoint 3 with the ball (indicated by the orange ball marker)

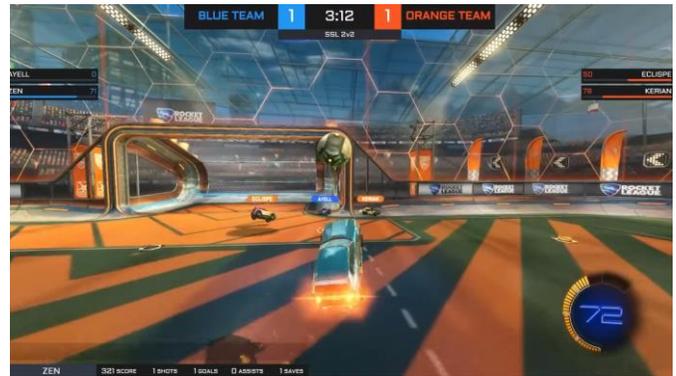

**Fig. 7.** A screenshot from a 2 vs 2 match in *Rocket League* shows cars competing to hit the ball toward the goal in a soccer-like arena. The ball is airborne near the goal as players control their cars to score or defend. The score and remaining time are displayed on the top of the user interface.

The observation space is derived directly from the game state. Relevant game entities are always observed, including the ball, opponents, allies, goals, checkpoints, and laps. For each entity, key-data points such as relative position, speed, and clear line-of-sight flags are observed. For players, flags indicating whether they are hurt, in air, or performing certain actions are included. The agent also observes its own position, state, and speed. To further enhance stability during training, Action Masking is used to eliminate infeasible actions. The reward structure is designed to incentivize game progress and teamwork, with scaled rewards for checkpoint completion and goals, complemented by penalties for opposing team successes to encourage competitive and cooperative behaviors. As an example, **Fig. 6** shows the cooperative and competitive behavior from a trained agent. Agents can strategically position themselves in offensive or defensive roles depending on the game state.

Additionally, agents are designed for multi-purpose functionality, operating across different game modes, including competitive and practice settings. In the mode "Training with Bots," the agent adjusts difficulty levels to accommodate various player skills, and in classic live matches, it seamlessly replaces human players who have disconnected, using lower difficulty models to maintain balance and ensure a smooth gameplay experience. Moreover, this system represents a practical step in game production, providing a scalable and efficient AI solution that enhances both gameplay and development processes.

### 4) *Rocket League*

*Rocket League* is a fast-paced vehicular soccer game where players control rocket-powered cars to score goals in an arena similar to a soccer field, combining elements of soccer strategy with driving mechanics. The game is primarily played in 2 vs. 2 or 3 vs. 3 team formats, where players coordinate with their teammates to maintain possession of the ball, defend their goal, and create scoring opportunities (**Fig. 7**). The core gameplay involves driving, jumping, boosting, and performing aerial maneuvers to outplay opponents, making each match a combination of skill and strategy.

Building on this interest in sports-like games for RL

research, *Rocket League* serves as an effective platform due to its combination of balanced gameplay, clearly defined competitive and cooperative interactions, and a structured, intuitive soccer-like environment. As discussed earlier, *Roller Champions* utilized the Unity ML-Agents Toolkit [57] for studies in MARL. Similarly, the same toolkit is utilized to adopt a two-agent approach for training AI agents in specialized tasks such as goalkeeping and striking in *Rocket League* [60]. This research illustrates the efficacy of sim-to-sim transfer. In the context of MARL, another work highlights the potential of *Rocket League* for developing team-based AI by leveraging its dynamic role-switching mechanics, where roles such as attacker, receiver, and defender are continuously reassigned based on game states, enabling agents to adapt and coordinate effectively in real-time [61]. The paper also examines the opportunities for advancing tactical decision-making processes, where agents analyze player positions, ball trajectories, and dynamic team formations to refine both offensive and defensive strategies. Furthermore, it explores the game's potential to facilitate human-bot collaboration through the development of more sophisticated real-time communication protocols that could enhance coordination among agents. The scalability of *Rocket League* to accommodate varying team sizes and configurations makes it an attractive environment for researching complex multi-agent interactions.

More recently, Lucy-SKG [62] was introduced as an AI agent developed to advance learning and performance in the *Rocket League* environment. Unlike earlier work that focused on specific tasks, such as training AI agents for specific roles like goalkeeper or striker, Lucy-SKG employs a more holistic approach by integrating novel methodologies that enhance both individual abilities and coordinated team-based play. Central to its design is the Kinesthetic Reward Combination (KRC) technique, which refines the reward signals to better represent complex cooperative and competitive in-game behaviors, such as maintaining optimal positioning and controlling the ball with precision in dynamic situations. Additionally, Lucy-SKG utilizes auxiliary tasks, such as state prediction and reward estimation, which act as supplementary objectives to regularize the learning process. These auxiliary tasks improve sample efficiency by guiding the agent to learn a broader range of representations about the game environment, ultimately facilitating more effective decision-making in multi-agent previously leading bots, Necto and Nexto, the research demonstrates that Lucy-SKG achieves superior learning efficiency and gameplay performance.

Lucy-SKG utilizes a portion of the state space available through RLGym [63], a python API to treat the game as an OpenAI Gym environment. It is structured as a triplet that includes a latent array containing player-specific information, a byte array representing key game objects to focus on, and a key padding mask to accommodate varying numbers of players. The action space consists of 90 discrete action combinations, encompassing various actions. At the core of the learning process, the agent uses KRC as an alternative to linear reward combinations designed to measure the utility of complex phenomena by creating a compound reward signal that reflects high-level state quality. The KRC balances multiple reward components, enabling the learning of complex skills such as aligning the ball toward the goal while maintaining a close distance, and offers potential for further generalization in future work. Additionally, to better facilitate learning in cooperative multi-agent settings, the reward function incorporates a reward distribution function:

$$R'_i = (1 - \tau) * R'_i + \tau * \overline{R}'_{team} - \overline{R}'_{opponent} \quad (15)$$

that considers team spirit that encourages coordination. Furthermore, to enhance learning efficiency and performance, Lucy-SKG has auxiliary neural architectures trained on reward prediction and state representation tasks. These auxiliary tasks are designed to improve the agent's understanding of the environment by predicting future rewards and representing states more effectively. The training is conducted in an on-policy fashion, integrating these tasks with the main learning objective to accelerate learning speed and overall performance. Finally, the agent's neural network architecture is based on a Perceiver model with cross-attention mechanisms, which leverages MLPs and Transformers [64] to process high-dimensional state inputs and learn representations that account for the interdependencies between agents. The architecture is trained using PPO. Through a series of ablation studies, the research demonstrates that the combination of KRC, auxiliary tasks, and PPO enhances the agent's learning speed and overall performance, establishing a new benchmark for game AI in *Rocket League's* multi-agent setting.

### B. First-Person Perspective Games

First-person perspective (FPP) games, a superset of first-person shooter (FPS) games, are among the most popular genres in video games. Unlike isometric or top-down view strategy games, where AI agents manage multiple entities, FPP games set the player or AI agents in a first-person view and operate as a single entity navigating a 3D virtual world. This immersion closely mirrors human sensory experiences, making FPP games particularly relevant for applications requiring real-time decision-making from a first-person viewpoint. The challenges of FPP games mostly come from their partially observable environments. Additionally, these games often have more interactions with the environment, such as opening doors or hiding behind covers, which add complexity to gameplay and agent learning processes. Moreover, the continuous action spaces and high-dimensional visual inputs characteristic of FPP games further elevate the difficulty. By tackling these complex gameplay and AI challenges, research in FPP games can drive innovation in creating more adaptive and intelligent agents, enhancing both AI performance in real-world applications and game experiences with more responsive, human-like game AI.

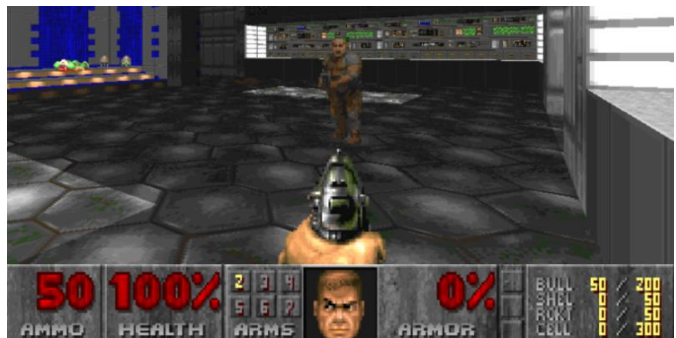

**Fig. 8**. *Doom's* first-person perspective view.

#### 1) *Doom & ViZDoom*

*Doom* is a well-known FPS video game released in 1993 that has significantly influenced the genre. In the game, players play the role of a space marine in first person view, navigating through a series of maze-like levels in military bases on Mars and Hell as shown in **Fig. 8**. The primary objective is to eliminate demonic enemies using a variety of weapons while collecting health packs, ammunition, and key cards to unlock new areas.

*Doom* also introduced a multiplayer mode, which includes cooperative gameplay and competitive deathmatch scenarios. In cooperative mode, players work together to complete levels by defeating enemies and achieving shared goals. In competitive deathmatch, players face off against each other in fast-paced battles to achieve the highest score by eliminating opponents. In recent years, a modified version ViZDoom [65] has become a prominent toolkit and environment for experimenting and developing RL algorithms in FPP environments. ViZDoom provides a 3D, FPP setup, allowing researchers to develop and test AI agents that play using only raw visual pixel input. This platform also supports a variety of scenarios, from basic target shooting to complex navigation and survival tasks. Notably, the multiplayer Deathmatch mode in ViZDoom has been popular in studying MARL [66].

The research in *Doom* and ViZDoom integrates CNNs with RL to manage observations similar to human vision. The action space in ViZDoom includes discrete actions such as moving, shooting, and navigating through levels. The reward strategy in these environments typically involves a combination of shaping positive and negative rewards, such as points for killing an opponent or penalties for self-damage.

One notable study is the development of the F1 agent [67], using Asynchronous Advantage Actor-Critic (A3C) [68] algorithm to train agents in the Deathmatch mode. The training process incorporated curriculum learning, starting with simple environments and progressively increasing the difficulty by introducing more challenging maps and stronger opponents. This approach resulted in the F1 agent achieving state-of-the-art performance, including winning Track 1 of the ViZDoom

AI Competition with a score 35% higher than the second-place competitor.

In conclusion, *Doom* and ViZDoom are excellent platforms to begin exploring MARL in FPP/FPS settings. Their relatively straightforward rules and moderate complexity make them ideal environments for developing and testing MARL algorithms before tackling more complex first-person view games.

### 2) *Minecraft*

As one of the most-played video games in the world, *Minecraft* has a giant open-world sandbox-style environment for players to explore, gather resources, craft tools, and build all kinds of complex structures (**Fig. 9**). It is also known for its multiplayer mode, where a world is shared among different players. In *Minecraft*, players begin by collecting basic materials such as wood and stone, which they use to create tools and shelters. As they progress, they can mine for rarer resources, build increasingly sophisticated structures, and engage in interactions like farming, trading, and combat.

The game's flexibility allows for a wide range of gameplay styles, making it a perfect platform for mixed-setting MARL research. In particular, cooperation happens when players working together to build large buildings, defend against in-game threats, or manage shared resources. On the other hand, competition is also a core aspect of multiplayer gameplay. Players or teams may compete to dominate resources, outperform one another in construction, or engage in direct combat. More interestingly, the roles are not always fixed, meaning players can switch between collaboration and competition, creating an unpredictable multiplayer experience and a mixed setting in MARL context.

Several research initiatives have utilized *Minecraft* to explore different aspects of AI, particularly in enhancing multi-agent interactions. One such effort involves using *Minecraft* as a simulated task environment (STE) to improve collaboration between human players and rule-based AI agents [69]. Another work done is the creation of BurlapCraft [70], a *Minecraft* mod developed to integrate with the BURLAP RL and planning library. BurlapCraft's integration with *Minecraft* enables AI agents to perform tasks such as navigation, block placement, and even language understanding, within a richly interactive 3D environment. Furthermore, the MineRL project [71] introduces a large-scale dataset of over 60 million state-action pairs of human demonstrations across various tasks in *Minecraft*. To capture the diversity of gameplay and player interactions, MineRL includes six tasks that present a range of research challenges, including open-world multi-agent interactions, long-term planning, vision, control, and navigation, as well as both explicit and implicit subtask hierarchies. These tasks are implemented as sequential decision-making environments within an existing *Minecraft* simulator. Additionally, the MineRL project features a novel platform and methodology for the continuous collection of human demonstrations. As users play on the publicly available MineRL server, their gameplay is recorded at the packet level, allowing for perfect reconstruction of each player's view and actions. This platform not only supports the ongoing expansion of the dataset with new tasks but also facilitates automatic annotation, making it a valuable resource for advancing AI research in complex, dynamic environments like *Minecraft*.

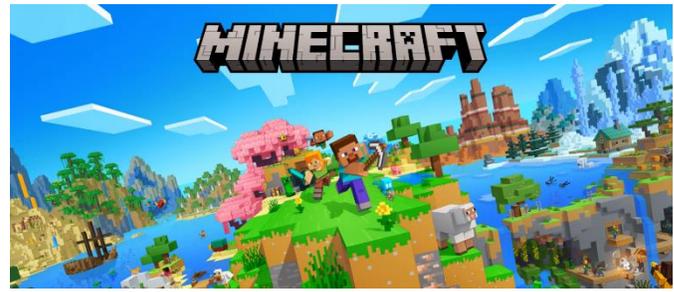

**Fig. 9.** An illustration of *Minecraft* game world.

Moreover, the developers of *Minecraft*, Microsoft, introduced Project Malmo [72], an AI experimentation platform built on top of the game to support a wide range of research in areas such as computer vision, RL/MARL, and robotics. It offers an abstraction layer and API on top of *Minecraft*, where multiple agents can interact with the environment by perceiving observations and rewards and taking actions in real-time. In 2017, Microsoft organized the Malmo Collaborative AI Challenge (MCAC) [73], aimed at advancing research in MARL through a collaborative mini-game within *Minecraft*. A notable champion of this challenge was *HogRider* [74]. *HogRider* was specifically designed to navigate the complexities of the *Pig Chase* mini-game, where agents either collaborate to catch a pig or deviate from cooperation to pursue individual gains. The developers employed a novel agent type hypothesis framework to identify and adapt to the behavior models of other agents, and a customized Q-learning. *HogRider*'s performance was exceptional, winning the challenge with a 13% higher mean score and 21.7% better variance-to-mean ratio than the second-best team. Additionally, it outperformed human players with a 28.1% higher mean score and a 29.6% reduction in the variance-to-mean ratio, demonstrating its superiority in both optimality and stability. Furthermore, the platform has also been employed to exploring MARL in MalmÖ (MARLÖ) Competition [75], which specifically challenges participants to develop MARL agents capable of generalizing across different mini-games and opponent types within *Minecraft*. The MARLÖ competition features three games: *Mob Chase*, a collaborative game where agents must work together to capture a mob; *Build Battle*, where teams compete to construct a specified structure; and *Treasure Hunt*, a mixed cooperative and competitive game involving resource collection and combat. These tasks are designed to engage both collaboration and competition among agents, highlighting the platform's potential to advance MARL research by fostering the development of versatile, general-purpose agents capable of learning across diverse scenarios. To sum up, TABLE III summarizes key features and focuses of these platforms.

*Minecraft* also presents an open-ended world without predefined win or lose conditions, closely simulating real-world scenarios. Despite the game's limitations, such as block-like graphics and simplified physics, its flexibility supports all three types of MARL interactions, making it an effective platform for studying complex multi-agent systems.

### 3) *Quake III Arena: Capture the Flag*

*Quake III Arena* is a multiplayer FPS game that was developed and released in 1999 by id Software. Aside from its regular gameplay, its Capture the Flag (CTF) mode has drawn attention in AI research in recent years. In this mode, two teams of players compete to capture the opponent's flag while defending their own within a 3D maze-like level (**Fig. 10**). The game's procedurally generated environments and adjustable map sizes offer a rich and flexible environment for studying strategic multi-agent cooperation and competition.

TABLE III: Research Platforms in *Minecraft*

| Platform | Focus | Key Features | Multi-Agent Interactions |
|---|---|---|---|
| STE | Human-AI Cooperation | Rule-based human-AI teaming | Rule-based collaboration |
| BURLAP | RL & Planning | sequential decision-making, navigation, block placement, language learning | hierarchical task management, MARL agent communication |
| MineRL | DRL Sample-Efficiency | Large-scale dataset of over 60 million state-action pairs | open-world multi-agent interactions from expert-level human demonstrations |
| MalmÖ | AI and AGI platform | computer vision, RL, robotics, multi-agent systems (MAS) | Real-time MAS for complex tasks, human interaction |

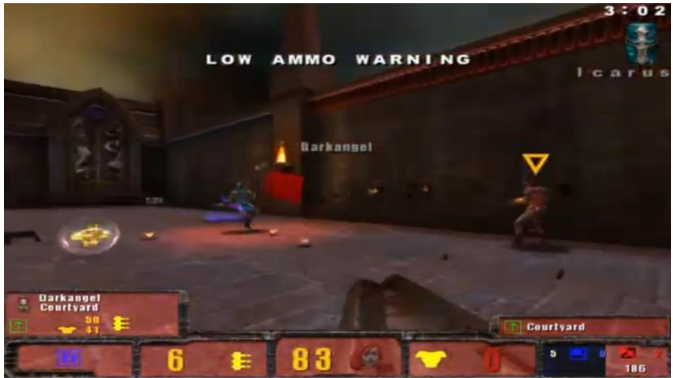

**Fig. 10.** *Quake III Team Arena - Capture The Flag*

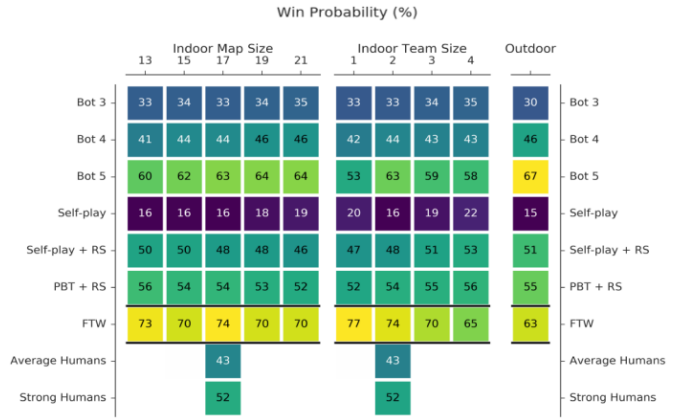

**Fig. 11.** The figure [76] shows the win probabilities of various AI agents and human players in CTF mode across different map sizes, team sizes, and environments. The FTW agents consistently achieve the highest win probabilities, surpassing both built-in scripted bots and human players, with over 70% win probability in most scenarios.

As one of the most notable works in this field, DeepMind trained MARL agents to play CTF using only pixel data and game points as inputs [76]. These agents learned directly from raw visual data, developing strategies for navigation, offense, and defense in a partially observable environment in real-time. Remarkably, the AI agents achieved performance levels that surpassed those of strong human players in competitive tournament settings.

In this context, their agents' observations in the CTF mode are closely modeled after those of human players, utilizing 84 x 84 pixel RGB frames processed through CNNs along with game points as input. The agent's policy $\pi_i$ is designed to maximize the probability of winning for its team $\{\pi_0, \pi_1, \ldots, \pi_{N/2-1}\}$, which is composed of $\pi_0$ itself and its teammates' policies $\pi_1, \ldots, \pi_{N/2-1}$, for a total of N players, defined as:

$$\mathbb{P}(\pi_0\text{'s team win} \mid \omega, (\pi_n)_{n=0}^{N-1})$$

$$= \mathbb{E}_{\omega \sim (\pi_n)_{n=0}^{N-1}} \left[ \left\{ \pi_0, \pi_1, \ldots, \pi_{\frac{N}{2}-1} \right\} \underset{>}{\bowtie} \left\{ \pi_{\frac{N}{2}}, \ldots, \pi_{N-1} \right\} \right]. \quad (16)$$

The winning operator $\underset{>}{\bowtie}$ returns 1 if the left team wins, 0 if it loses, and resolves ties randomly. $\omega$ represents the specific map instance and random seeds, which are stochastic in learning and testing. The binary outcome of win/lose as a reward is insufficient for effective learning due to sparse and delayed rewards. To address this, more frequent rewards are introduced that correspond to specific in-game events, such as capturing the flag, picking up the flag, or having a teammate capture the flag etc. These rewards can be utilized directly in reward shaping, or they can be transformed into a reward signal through a learned transformation function, forming the foundation of its novel For-The-Win (FTW) agent architecture. The architecture consists of two LSTM networks operating at distinct timescales: the fast LSTM processes inputs such as pixel observations, previous actions, and rewards at each environment step, while the slow LSTM updates at a lower frequency, capturing long-term temporal dependencies and facilitating strategic planning. In addition, the fast LSTM generates a variational posterior distribution, incorporating new observations and prior knowledge from the slow LSTM, which generates a prior distribution on the latent variable. Moreover, the architecture is augmented with an external Differentiable Neural Computer (DNC) memory module, which enhances the agent's ability to store and retrieve past experiences, mimicking episodic memory functions. Lastly, the FTW agent uses an extensive action space consisting of 540 distinct actions, generated by combining elements from six independent action dimensions.

Optimization within the FTW architecture is performed using a two-tier approach. The first tier, or inner optimization, focuses on maximizing the agents' expected future discounted internal rewards. Complementing this is the second tier, or outer optimization that is managed through Population-Based Training (PBT). PBT is an online evolutionary process, adapting internal rewards and hyperparameters while performing model selection. In this process, agents that underperform are systematically replaced with mutated versions of better-performing agents, ensuring that the population evolves towards higher performance. The effectiveness of PBT is further enhanced by self-play and a distributed training architecture featuring an actor-learner structure, with 1920 parallel arena processes facilitating large-scale, concurrent training.

The agents demonstrated superior performance across various scenarios. As a result, the FTW agents consistently outperformed human players, capturing an average of 16 more flags per game on procedurally generated maps that neither the agents nor the humans had previously encountered. Even when human players were paired with FTW agents, the human-agent teams had only a 5% win probability against a team of two FTW agents. In a targeted test, where professional game testers had six hours to devise exploitative strategies against the FTW agents on a complex map, the humans could only achieve a 25% win rate. An overview of the test results is shown in **Fig. 11**. The evaluation also highlighted significant differences between the agents and human players in terms of reaction times and tagging accuracy. The FTW agents had a reaction time of 258 milliseconds, compared to 559 milliseconds for humans, and a tagging accuracy of 80%, substantially higher than the 48% accuracy of human players. Even when the FTW agents' tagging accuracy was deliberately reduced to match human levels, they still maintained a higher win probability. These results showcase the effectiveness of the FTW architecture and training methodologies, with the agents consistently outperforming both human players and existing AI benchmarks in the CTF mode.

### C. RTS and MOBA Games

Real-time strategy (RTS) is a subgenre of strategy games, but instead of playing in turns, RTS games allow all players to play simultaneously in real-time. The genre is defined by resource management, base building, and large-scale tactical combat, typically with a top-down or isometric camera view that gives players a broad perspective of the battlefield. Players control multiple worker or soldier units and structures, making real-time decisions to gather resources, build bases, and engage in strategic combat against opponents. These games require quick thinking and efficient management of various tasks simultaneously. Building on this foundation, the Multiplayer Online Battle Arena (MOBA) subgenre emerged, thriving as one of the most popular gaming formats. MOBA games simplify individual player actions by focusing on controlling one single character, and shift the emphasis to team collaboration and strategic coordination rather than multitasking from RTS games. The camera view remains similar, often top-down, but the gameplay centers on working with teammates to achieve certain objectives, manage map control, and outplay the opposing team. Both RTS and MOBA games involve partially observable environments with fog of war, long-term strategic planning, high APM, and coordinated teamwork to influence the game's outcome over long periods.

#### 1) *StarCraft II*

As one of the most well-known RTS games, *StarCraft II* was created by Blizzard Entertainment as a sequel to *StarCraft*. Players choose one of three races, gather resources to progress, build structures, and eventually defeat other players or AI agents with soldier units from their chosen race. Like most RTS games, *StarCraft II* is an imperfect information game where players must rely on a combination of a local camera that limits their view to a specific area and a minimap (**Fig. 12**) that provides only a high-level overview of the battlefield. Additionally, the fog of war obscures unobserved regions of the map, making active exploration essential to gathering information about the opponent's state.

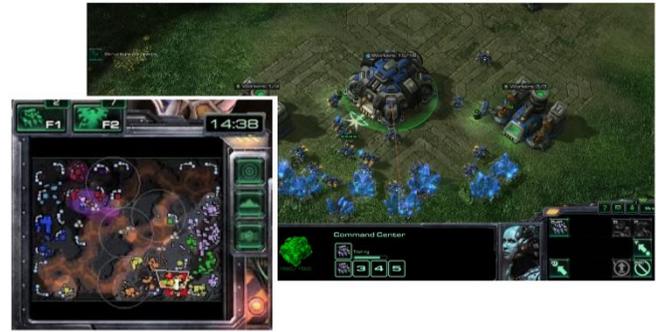

**Fig. 12.** *StarCraft II* Minimap: it provides an overview of the entire battlefield, allowing players to monitor key locations, track allied and enemy movements, and quickly navigate between different areas. Players can also ping the minimap to communicate with their team, improving coordination.

In 2017, DeepMind, in collaboration with Blizzard, introduced the StarCraft II Learning Environment (SC2LE), a platform based on *StarCraft II* to support research on RL with multi-agent support [45]. Using this platform, DeepMind developed AlphaStar in 2019 [23], which reached a significant milestone by defeating professional human players in *StarCraft II* tournaments. It was rated at Grandmaster level for all three races and above 99.8% of officially ranked human players.

The state space in *StarCraft II* is vast, with each game consisting of tens of thousands of time-steps and thousands of actions, selected in real-time throughout approximately ten minutes of gameplay. Every second in the game, there are typically 16 time-steps. At each step AlphaStar receives an observation that includes a list of all observable units and their attributes, although this information is imperfect as it includes only opponent units that are visible to the player's own units. The action space in *StarCraft II* is highly structured, resulting in approximately $10^{26}$ possible choices at each step.

AlphaStar's architecture involved a Transformer [64] based neural network to process the game's extensive state space. Its training process included supervised learning from human expert games, followed by RL through self-play with a novel "League Training" approach. Initially, it was trained on a dataset of 971,000 replays from the top 22% of players, with a fine-tuning phase using games with Matchmaking Rating (MMR) above 6,200 while the highest is typically around 7,000, achieving a supervised learning rating above 84% of human players. The core RL algorithm employed was a variant of PPO for its high-dimensional action spaces.

AlphaStar's league training approach is designed to address the exploration-exploitation dilemma [77] in *StarCraft II*. The training involves creating different leagues of agents that compete against each other through a multi-faceted system of main agents, main exploiters, and league exploiters. Main agents continuously improve by playing against both current and past versions of themselves by using Prioritized Fictitious Self-Play (PFSP) to focus on opponents that pose the greatest challenge. Main exploiters are specifically trained to identify and exploit weaknesses in the main agents to improve their adaptability and resilience. League exploiters, on the other hand, target weaknesses in the overall league. This structure can might regress or get stuck in loops of non-transitive strategies. This is particularly relevant in *StarCraft II*, which has a similar setup with three races that counter each other. Additionally, AlphaStar's use of human data for initial policy training and strategic diversity, combined with RL techniques like V-trace and upgoing policy updates (UPGO), fosters continuous improvement and adaptation within the league training system. These techniques have been driving ongoing research and innovation in MARL, inspiring new approaches to training AI agents in other real-time environments.

In addition to the advancements made by AlphaStar, other research efforts have also contributed to the development of MARL in *StarCraft II*. A grid-wise control architecture has been presented to address the challenges associated with managing varying numbers of agents in spatial grid environments [79]. This architecture enhances multi-agent collaboration through a convolutional encoder-decoder network that enables scalable and flexible coordination among agents. In addition, a noval hierarchical control framework for MARL in RTS games improves multi-agent coordination by separating decision-making into macro-strategies at the high level and micro-actions at the lower level [80]. This division allows for the execution of hierarchical, coordinated strategies that integrate strategic planning with tactical execution.

Although the *StarCraft* series is popular in various multi-agent research studies, the majority remains focused on 1 vs. 1 competitive gameplay even though *StarCraft II* supports up to 12 players and can be trained using league systems or many vs. many configurations. The extensive micromanagement required to play *StarCraft II* makes the game both intense and for many players. On the other hand, MOBA games like *Dota 2* have thrived. These games emphasize 5 vs. 5 teamwork over individual player skills with much simpler control schemes, single-unit management and single-resource management. This shift not only captured the attention of a broader player base but also created a better environment for MARL research, where the focus is on team collaboration rather than just individual skills or APM.

### 2) OpenAI Five for Dota 2

*Defense of the Ancients 2* (Dota 2), developed by Valve Corporation, is a leading title in the popular MOBA genre. The game features two teams of five players, each controlling a "hero" with unique skills, complemented by various equitable items that provide passive enhancements, functionalities, and active abilities. The primary objective is to destroy the opposing team's Ancient, the primary structure in their base. An example of player's observation in *Dota 2* is shown in **Fig. 13**, along with some of critical gameplay elements.

The complexity of *Dota 2*, characterized by its large and partially observable environment, long game duration, extremely sparse and delayed rewards, vast and continuous state and action spaces, demands real-time strategic planning and precise team coordination. Unlike traditional RTS games like *StarCraft II*, which focus mainly on 1 vs. 1 gameplay and micro-operation, *Dota 2* emphasizes 5 vs. 5 teamwork, where success depends on team coordination, macro strategies and sequential decision-making. This shift to multi-agent collaboration has not only resonated with the player community but has also positioned *Dota 2* as a premier platform for advancing research in MARL. The potential of these technologies was demonstrated in 2018 when OpenAI Five achieved a groundbreaking victory over the world's best human players, Team OG, at The International 2018.

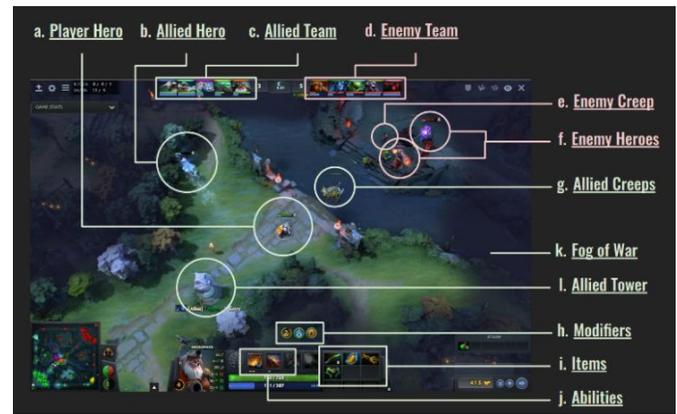

**Fig. 13.** *Dota 2* Human Observation Space [24]

The OpenAI Five [24] system employed a large-scale DRL approach to train AI agents to play *Dota 2* at a superhuman level. The system tackled a high-dimensional state space of approximately 16,000 continuous and categorical variables per time-step, while the action space required the model to select among 8,000 to 80,000 possible actions, depending on the hero. To manage, the training process relied on a RNN with approximately 159 million parameters, centered around a single-layer LSTM with 4096 units, which constituted 84% of the model's parameters. Each hero in a team was controlled by a replica of this network.

To mitigate the challenges posed by *Dota 2*'s complexity, certain game mechanics were controlled by hand-scripted logic rather than the policy. This included the order in which heroes purchased items and abilities, the control of the unique courier unit, and the management of items heroes kept in reserve. Additionally, the system was restricted to a subset of 17 heroes out of the 117 available in the full game, and it excluded support for items that allow players to temporarily control multiple units simultaneously, reducing the technical complexity and computational demands.

The training process also utilized PPO, enhanced with Generalized Advantage Estimation (GAE) to stabilize and accelerate learning. The policy optimization was conducted on a distributed system, employing up to 1536 GPUs and achieving a batch size of nearly 3 million time-steps per update. Over approximately 10 months, OpenAI Five played 45,000 years of *Dota 2* games using 128,000 CPU cores and 256 GPUs. To address the challenge of long-term credit assignment, given *Dota 2*'s lengthy episodes and sparse rewards, OpenAI Five incorporated a novel reward structure that symmetrized rewards by subtracting the opponent's rewards from the team's, alongside a dynamic opponent sampling system to maintain robustness and prevent strategy collapse. Furthermore, OpenAI Five overcame significant challenges by implementing a continuous development process called "surgery," which allowed for iterative improvements without restarting training.

OpenAI Five also effectively addressed several collaboration challenges such as lane assignments, hero lineup diversity, and team incentives. Initially, agents struggled with proper lane assignments, often clustering together and undermining long-term strategies. A penalty system was introduced to reinforce lane discipline, while randomized hero lineups during training ensured the robustness across different team compositions. Central to its MARL approach was the "Team Spirit" hyperparameter, which managed the credit assignment problem among the five agents. By adjusting how rewards were shared among teammates, Team Spirit balanced individual and collective incentives. During early training, a lower Team Spirit reduced gradient variance, helping agents refine their mechanical and tactical abilities. As training progressed, increasing Team Spirit shifted the focus toward optimizing actions for overall team success.

As a result, OpenAI Five's success demonstrates the potential of MARL in complex MOBA games. It competed against 3,193 unique teams in 7,257 total games with a 99.4% win rate, including a 2-0 victory over Team OG at The International 2018. These outcomes highlight OpenAI Five's superhuman capabilities and provide valuable insights for scenarios requiring large-scale planning, coordination, and decision-making across multiple agents.

### 3) *Honor of Kings*

Similarly, *Honor of Kings* is a MOBA game played on mobile phones (**Fig. 14**). It focuses on 5 vs. 5 team coordination to destroy the opponents' Crystal just like *Dota 2*. While *Honor of Kings* offers a classic MOBA environment, it has further reduced complexity compared to *Dota 2*, with a smaller map, fewer hero skills, fewer active items, and typically shorter game durations. Despite these differences, the game retains a high level of strategic depth, making it another ideal gaming platform for MARL.

Notably, in 2020, Tencent AI Lab developed an AI system that achieved superhuman performance in *Honor of Kings*, winning 95.2% of 42 matches against professional esports players and 97.7% of 642,047 matches against top-ranked human players [81]. Unlike OpenAI Five, which was limited to supporting only 17 heroes out of 117 due to the complexity of training across a broad hero pool, Tencent's system successfully managed a pool of 40 heroes, representing a step toward fully capable MOBA AI.

The AI system for *Honor of Kings* employs a combination of novel and existing DRL techniques. The state space consists of 9,227 scalar features and 6 channels of spatial features with a resolution of 6x17x17, while the action space includes control actions for hero movements, attacks, and use of skills. A significant implementation is the unified actor-critic network architecture, which captures the playing mechanics and actions of 40 unique heroes. The architecture integrates convolutional layers for spatial features and fully connected layers for scalar features, combined with an LSTM to manage temporal dependencies. The training process involves a curriculum self-play learning (CSPL) with of three phases: fixed-lineup training, multi-teacher policy distillation, and random-pick training. Initially, fixed lineups are trained separately to create teacher models. These models, with 9 million parameters each, are then distilled into a single student model with 17 million parameters. The final phase involves training the student model with randomly picked hero lineups from a pool of 40 heroes.

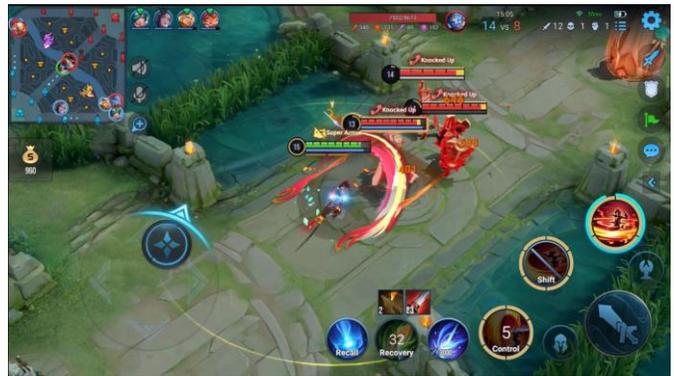

**Fig. 14.** *Honor of Kings* player view: player controls a single hero unit shown in the middle with green health bars. A minimap is also available to the player in the top-left corner of the screen.

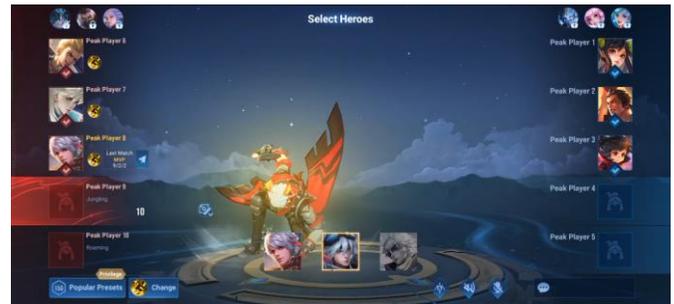

**Fig. 15.** Hero selection (drafting) phase in *Honor of Kings*: players choose their heroes before the start of a match. Each player selects a hero from a pool of available characters in turns.

In all MOBA games, there is a phase called "drafting," where both teams strategically select their heroes in turns before the game starts (**Fig. 15**). The choice of heroes is extremely important to the team's overall strategy and success as each hero has unique abilities and strengths that can complement or counteract other heroes. Selecting the right combination of heroes can significantly impact the outcome of the game. In traditional MOBA play, this phase requires deep understanding and foresight, as teams must anticipate the opponent's picks and choose heroes that provide synergies with their own team while countering the opposing team's lineup. This drafting phase further complicates the MARL learning process, particularly due to the extensive combinatorial space. Given a hero pool consisting of 40 options, the number of possible combinations is $C_{40}^{10} \times C_{40}^{5}$, which exceeds $10^{11}$. Expanding this to a full game of *Honor of Kings* with 113 heroes, the possible lineup combinations increase dramatically to $1.56 \times 10^{16}$ [82]. This vast number of combinations makes a complete tree search method, such as the Minimax algorithm used in OpenAI Five, computationally infeasible [83].

To address the challenges associated with drafting in MOBA games, the AI system in *Honor of Kings* implements MCTS to simulate various possible hero combinations and their outcomes. The AI system uses a value network trained on 100 million samples to predict the win rates for different hero combinations, allowing it to evaluate and prioritize the most effective hero combinations. In addition, the training and execution of this sophisticated drafting strategy are supported by substantial computational resources, including 320 GPUs

and 35,000 CPUs.

The AI system in *Honor of Kings* marks a substantial advancement MARL by effectively addressing the challenges of hero selection in MOBA games. While previous systems, such as OpenAI Five, were constrained by the complexity of managing a limited pool of 17 heroes [24], *Honor of Kings* expanded this capability to include 40 heroes. This development not only advances the capabilities of AI in gaming systems but also offers insights into the broader applications of MARL in other domains.

So far, we have reviewed the applications of MARL in games with many participants across various genres, following the order of game complexity introduced earlier, and emphasizing implementation details, agent interactions and performance. To further summarize, TABLE IV provides an overview of these findings. The next section will delve into a discussion of the current implications and the exploration of future directions.

## VIII. DISCUSSION

While MARL has demonstrated remarkable success in research settings, its integration into video game industry remains limited. This section discusses the key barriers to industrial adoption, including production constraints, design priorities, and control requirements that differ from academic assumptions. It also explores how these challenges affect the application of MARL in commercial games and outlines potential solutions that may help bridge the gap between research and practice.

### A. Industry and Academia Gap

While MARL has made significant progress in recent years with projects like AlphaStar and OpenAI Five, the majority of game industry has gone on a different path that focuses on Game AI, a term referring to handcrafted, predictable, and manageable AI behaviors. In 2014, a review highlighted that RTS games rely on finite state machines and behavior trees rather than advanced RL techniques [10]. This is still true for most of game AI in the industry today. This divergence reflects the difficulty of integrating academic AI techniques into practical game development workflows, where predictability and designer control take precedence over cutting-edge but unpredictable AI behaviors. Additionally, advanced AI techniques add significant cost to game development. On top of development budget, lack of adaptability introduces further expenses. These AI systems are usually developed for specific game environments, but game developers typically need them to work across multiple projects. One direction to bridge this gap is the development of generalist agents capable of handling multiple roles, modes or games. Recent work in environments like *Rocket League* and *Minecraft* shows that agents can adapt to dynamic roles or generalize across diverse modes.

### B. Superhuman AI vs. Human-like AI

Most MARL research projects, such as AlphaStar and OpenAI Five, measure success by how much AI surpasses human performance. However, for video games, superhuman

AI is often undesirable, as unbeatable opponents destroy the enjoyment and mental challenge for human players. Instead, human-like AI can provide engaging interactions through effective cooperation and fair competition. AI that can adapt to player skill levels and offer a variety of interaction for both companionship and rivalry would not only enhance game experience but also contribute to broader AI research such as AGI by developing agents that better understand and replicate human behavior.

### C. Creating Designer Centric RL

The game industry prioritizes AI that enhances player experience over purely optimal solutions. Designers require control over AI behavior to align with the game's narrative and aesthetic, balancing autonomy with predictability. While behavior trees and finite state machines offer control, they limit emergent behavior. In contrast, RL allows for adaptive agents but is often too rigid to meet design constraints. To bridge this gap, techniques like preference learning and potential-based reward shaping have been proposed [84].

### D. Applying AI to Other Game Genres

MARL research has demonstrated superhuman performance in genres like Sports, FPS, MOBAs and RTS. However, other genres present completely different challenges beyond directly competing with players. For example, in turn-based strategy games such as *Civilization VI*, AI is critical to gameplay where it needs to run a human civilization as the leader managing a variety of game systems including diplomacy, economics, country construction, military strategy, and resource management. In addition, each of the system has interactions among players. Scaling RL to such genres will boost innovations in hierarchical RL and multi-objective optimization, contributing to smarter and more capable AI instead of superhuman AI.

### E. Accessibility for Small Game Studios

MARL is currently limited to large studios with access to extensive resources. To push MARL in game industry, accessible toolkits like Unity ML-Agents Toolkit and cost-efficient training methods are crucial. Solutions such as pre-trained models, cloud-based MARL platforms, and general-purpose algorithms will enable smaller studios to implement advanced AI and utilize the technology to improve gaming experience. Lowering the bar to entry will not only foster innovation but also diversify the types of games developed with AI, and ultimately revolutionize the industry.

## IX. CONCLUSION

We review the applications of MARL across two-agent and multi-agent games in popular genres such as sports, FPS, RTS, and MOBA. While these implementations have pushed AI forward, the game industry still faces challenges in adopting RL or MARL as developers prioritize control, predictability, and budget. With the development of AI and other technologies in recent years, there is increasing demand for adaptive, human-like agents that can enhance player experiences while balancing cost-efficiency. Additionally, making MARL more accessible to researchers and developers will unlock its potential across diverse game genres. Despite the current challenges, video games offer massive opportunities not only within the gaming industry, but also continue to be the frontier for AI innovation and a platform for real-world applications.

TABLE IV: MARL Overview in Multi-Agent Games

| | Game | State Space (per state) | # of Actions (per step) | Reward Strategy | NN Architecture | Training | Multi-agent Focuses | Research Highlights |
|---|---|---|---|---|---|---|---|---|
| **Competitive and Sports Games** | **3v3 Snake** | 12×10×20 matrix, | 4 | zero-sum condition / rewards for survival / territory control | 8 Residual Blocks with 2x3x3 Conv layers | CTDE, Distributed PPO, 20 actors, 1 learner | territory control / shared rewards / rule-based team strategies | rule-enhanced MARL with territory matrix and masked illegal actions. |
| | **GRF** | 1280×720 (RGB) / 4×72×96 (SMM) / 115 floats. | 19 | SCORING (+1/-1 for goals), CHECKPOINT (+0.1 for field progression) | CNN, LSTM | IMPALA / PPO / Ape-X DQN | teamwork / counter-strategies / role-specific training | *Football Academy* with varied difficulty and scenarios. |
| | **Roller Champions** | 78 game entities | 9 | Team-based / checkpoints / goal scoring | MLP (3 layers x 512 neurons). | CTDE, PPO, Self-play, 3–15 instances. | self-assigned roles / shared team rewards / team strategic positioning | self-assigned roles, dynamic difficulty |
| | **Rocket League (Lucy-SKG)** | 3 arrays of game states | 90 options | KRC | Perceiver, MLP, transformer. | CTDE, PPO, auxiliary tasks | Cooperative learning, | Outperforms Necto/Nexto; benchmarks for Rocket League AI. |
| **FPP Games** | **ViZDoom** | 320×240 RGB | 8 | Task-based | CNN | DQN / SARSA / A3C, DTDE | up to 16 agents / team-based / deathmatch settings | custom scenarios / synchronous and asynchronous modes / vision-based RL with raw visual input only |
| | **Minecraft** | (Varies by Platforms – STE, BURLAP, MineRL, Malmo, etc) | | | | | | |
| | **Quake III Arena: CTF** | 84×84 RGB | 540 | Team-based Reward Shaping | CNN, Hierarchical RNN of 2 LSTMs | CTDE, Actor-Learner, Distributed PBT | Incentive for coordination and evolution. | PBT, FTW Agent, Temporal Hierarchy |
| **RTS and MOBA Games** | **StarCraft II (AlphaStar)** | Camera view of all visible units and their attributes, 256×256 grid | $10^{26}$ | Outcome-based (Win/Loss/Draw), pseudo-rewards | self-attention, scatter connections, LSTM, auto-regressive policy, transformer, pointer networks | CTDE, Supervised Learning, TD($\lambda$), V-Trace, UPGO, off-policy corrections. | League training with PFSP | League training with PFSP, beat 99.8% of human players |
| | **Dota 2** | 16,000 inputs on game state | 8,000 - 80,000 | Game outcomes (win/loss), additional rewards shaped by in-game events. | 4096-unit LSTM | CTDE, PPO with GAE | shared information, Incentive for team coordination | "Surgery" technique, large-scale distributed self-play, beat world champions |
| | **Honor of Kings** | 9,227 scalar features + 6×17×17 spatial features | 10 options | Outcome-based (Win/Loss), additional rewards shaped by in-game events. | CNN, MLP, LSTM | CTDE, Actor-Critic, Dual-clip PPO | CSPL, multi-teacher policy distillation, strategic team hero selections (drafting), | MCTS for hero selections (drafting), 95.2% win-rate over 42 matches against professionals |

# REFERENCES

[1] "Video games remain America's favorite pastime with more than 212 million Americans playing regularly," *Entertainment Software Association*, Washington, DC, USA, Press Release, Jul. 10, 2023. [Online]. Available: https://www.theesa.com/video-games-remain-americas-favorite-pastime-with-more-than-212-million-americans-playing-regularly/

[2] Video Game Market Size, Share & Trends Analysis Report by Device (Console, Mobile, Computer), by Type (Online, Offline), by Region (Asia Pacific, North America, Europe), and Segment Forecasts, 2023–2030, Horizon Databook, San Francisco, CA, USA, Rep. GVR-4-68038-527-4, 2023. [Online]. Available: https://www.grandviewresearch.com/industry-analysis/video-game-market

[3] "2024 essential facts about the U.S. video game industry," *the ESA*, 2024. [Online]. Available: https://www.theesa.com/resources/essential-facts-about-the-us-video-game-industry/2024-data/

[4] R. Redheffer, "A machine for playing the game Nim," *Am. Math. Mon.*, vol. 55, no. 6, pp. 343–349, Jun. 1948.,

[5] A. J. Champandard, "Understanding behavior trees," *AiGameDev.com*, no. 6, 2007.

[6] R. S. Sutton and A. G. Barto, *Reinforcement Learning: An Introduction*. Cambridge, MA, USA: MIT Press, 1998.

[7] J. Manslow, "Using reinforcement learning to solve AI control problems," in *AI Game Programming Wisdom 2*, S. Rabin, Ed. Hingham, MA, USA: Charles River Media, 2004.

[8] A. Shantia, E. Begue, and M. Wiering, "Connectionist reinforcement learning for intelligent unit micromanagement in StarCraft," in *Proc. Int. Joint Conf. Neural Netw.*, San Jose, CA, USA, 2011, pp. 1794–1801, doi: 10.1109/IJCNN.2011.6033442.

[9] E. Pagalyte, M. Mancini, and L. Climent, "Go with the flow: Reinforcement learning in turn-based battle video games," in *Proc. 20th ACM Int. Conf. Intell. Virtual Agents (IVA)*, Oct. 2020, pp. 1–8, doi: 10.1145/3383652.3423868.

[10] G. Robertson and I. Watson, "A review of real-time strategy game AI," *AI Mag.*, vol. 35, no. 4, pp. 75–104, Dec. 2014.,

[11] V. Mnih, K. Kavukcuoglu, D. Silver, A. A. Rusu, J. Veness, M. G. Bellemare, *et al.*, "Human-level control through deep reinforcement learning," *Nature*, vol. 518, no. 7540, pp. 529–533, Feb. 26 2015.,

[12] G. Lample and D. S. Chaplot, "Playing FPS games with deep reinforcement learning," in *Proc. AAAI Conf. Artif. Intell.*, vol. 31, no. 1, Feb. 2017, doi: 10.1609/aaai.v31i1.10827.

[13] E. Alonso, M. Peter, D. Goumard, and J. Romoff, "Deep reinforcement learning for navigation in AAA video games," in *Proc. 30th Int. Joint Conf. Artif. Intell. (IJCAI)*, Z.-H. Zhou, Ed., Montreal, QC, Canada, Aug. 2021, pp. 2133–2139. doi: 10.24963/ijcai.2021/294.

[14] S. J. Russell and P. Norvig, *Artificial Intelligence: A Modern Approach*, 4th ed. Boston, MA, USA: Pearson, 2018.

[15] J. Schrittwieser, I. Antonoglou, T. Hubert, K. Simonyan, L. Sifre, S. Schmitt, *et al.*, "Mastering Atari, Go, chess and shogi by planning with a learned model," *Nature*, vol. 588, no. 7839, pp. 604–609, Dec. 2020.,

[16] R. Imamura, T. Seno, K. Kawamoto, and M. Spranger, "Expert human-level driving in Gran Turismo Sport using deep reinforcement learning with image-based representation," *arXiv:2111.06449*, 2021.

[17] F. Fuchs, Y. Song, E. Kaufmann, D. Scaramuzza, and P. Duerr, "Super-human performance in Gran Turismo Sport using deep reinforcement learning," *IEEE Robot. Autom. Lett.*, vol. 6, no. 2, pp. 1–8, Apr. 2021.,

[18] J. Gillberg, J. Bergdahl, A. Sestini, A. Eakins, and L. Gisslén, "Technical challenges of deploying reinforcement learning agents for game testing in AAA games," in *Proc. IEEE Conf. Games (CoG)*, Boston, MA, USA, 2023, pp. 1–8, doi: 10.1109/CoG57401.2023.10333194.

[19] G. Tesauro, "TD-Gammon, a self-teaching backgammon program, achieves master-level play," *Neural Comput.*, vol. 6, no. 2, pp. 215–219, Mar. 1994.

[20] D. Silver, A. Huang, C. J. Maddison, A. Guez, L. Sifre, G. van den Driessche, *et al.*, "Mastering the game of Go with deep neural networks and tree search," *Nature*, vol. 529, no. 7587, pp. 484–489, Jan. 28 2016.,

[21] D. Silver, J. Schrittwieser, K. Simonyan, I. Antonoglou, A. Huang, A. Guez, *et al.*, "Mastering the game of Go without human knowledge," *Nature*, vol. 550, no. 7676, pp. 354–359, Oct. 18 2017.,

[22] D. Silver, T. Hubert, J. Schrittwieser, I. Antonoglou, M. Lai, A. Guez, *et al.*, "A general reinforcement learning algorithm that masters chess, shogi, and Go through self-play," *Science*, vol. 362, no. 6419, pp. 1140–1144, Dec. 7 2018.,

[23] O. Vinyals, I. Babuschkin, W. M. Czarnecki, M. Mathieu, A. Dudzik, J. Chung, *et al.*, "Grandmaster level in StarCraft II using multi-agent reinforcement learning," *Nature*, vol. 575, no. 7782, pp. 350–354, Nov. 2019.,

[24] OpenAI *et al.*, "Dota 2 with large scale deep reinforcement learning," *arXiv:1912.06680 [cs, stat]*, Dec. 2019.

[25] L. Busoniu, R. Babuska, and B. De Schutter, "A comprehensive survey of multiagent reinforcement learning," *IEEE Trans. Syst., Man, Cybern., Part C (Appl. Rev.)*, vol. 38, no. 2, pp. 156–172, Mar. 2008, doi: 10.1109/TSMCC.2007.913919.

[26] K. Zhang, Z. Yang, and T. Başar, "Multi-agent reinforcement learning: A selective overview of theories and algorithms," in *Handbook of Reinforcement Learning and Control*, pp. 321–384, 2021, doi: 10.1007/978-3-030-60990-0_12.

[27] T. T. Nguyen, N. D. Nguyen, and S. Nahavandi, "Deep reinforcement learning for multiagent systems: A review of challenges, solutions, and applications," *IEEE Trans. Cybern.*, vol. 50, no. 9, pp. 3826–3839, Sep. 2020.,

[28] L. Canese, G. C. Cardarilli, L. Di Nunzio, R. Fazzolari, D. Giardino, M. Re, *et al.*, "Multi-agent reinforcement learning: A review of challenges and applications," *Appl. Sci. (Basel)*, vol. 11, no. 11, p. 4948, May 2021.,

[29] N. Justesen, P. Bontrager, J. Togelius, and S. Risi, "Deep learning for video game playing," *IEEE Trans. Games*, vol. 12, no. 1, pp. 1–20, Mar. 2020.,

[30] K. Shao, Z. Tang, Y. Zhu, N. Li, and D. Zhao, "A survey of deep reinforcement learning in video games," *arXiv:1912.10944 [cs]*, Dec. 2019. [Online]. Available: https://arxiv.org/abs/1912.10944

[31] K. Souchleris, G. K. Sidiropoulos, and G. A. Papakostas, "Reinforcement learning in game industry—Review, prospects and challenges," *Appl. Sci. (Basel)*, vol. 13, no. 4, p. 2441, Feb. 2023.,

[32] C. J. C. H. Watkins and P. Dayan, "Q-learning," *Mach. Learn.*, vol. 8, no. 3–4, pp. 279–292, 1992.

[33] R. S. Sutton, D. McAllester, S. Singh, and Y. Mansour, "Policy gradient methods for reinforcement learning with function approximation," in *Proc. 12th Int. Conf. Neural Inf. Process. Syst. (NIPS'99)*. Cambridge, MA, USA: MIT Press, 1999, pp. 1057–1063.

[34] M. Wooldridge, *An Introduction to MultiAgent Systems*, 2nd ed. Chichester, U.K.: Wiley, 2009

[35] K. Dill and C. Dragert, "Modular AI," in *Game AI Pro 3: Collected Wisdom of Game AI Professionals*, S. Rabin, Ed., Boca Raton, FL, USA: CRC Press, 2017.

[36] S. Hochreiter and J. Schmidhuber, "Long short-term memory," *Neural Comput.*, vol. 9, no. 8, pp. 1735–1780, Nov. 15 1997.

[37] M. Hausknecht and P. Stone, "Deep recurrent Q-learning for partially observable MDPs," in *Proc. AAAI Fall Symp. Ser.*, 2015.

[38] M. Hessel *et al.*, "Rainbow: Combining improvements in deep reinforcement learning," in *Proc. 32nd AAAI Conf. Artif. Intell. (AAAI'18)*, New Orleans, LA, USA, 2018, pp. 3215–3222.

[39] V. R. Konda and J. N. Tsitsiklis, "Actor-critic algorithms," in *Adv. Neural Inf. Process. Syst. (NIPS)*, 2000, pp. 1008–1014.

[40] T. P. Lillicrap, J. J. Hunt, A. Pritzel, N. Heess, T. Erez, Y. Tassa, D. Silver, and D. Wierstra, "Continuous control with deep reinforcement learning," *arXiv preprint* arXiv:1509.02971, 2019. [Online]. Available: https://arxiv.org/abs/1509.02971

[41] J. Schulman, F. Wolski, P. Dhariwal, A. Radford, and O. Klimov, "Proximal policy optimization algorithms," *arXiv preprint* arXiv:1707.06347, 2017.

[42] Y. Shoham and K. Leyton-Brown, *Multiagent Systems: Algorithmic, Game-Theoretic, and Logical Foundations*. New York, NY, USA: Cambridge Univ. Press, 2008. doi.org/10.1017/CBO9780511811654.

[43] M. L. Littman, "Markov games as a framework for multi-agent reinforcement learning," in *Proc. 11th Int. Conf. Mach. Learn. (ICML'94)*, San Francisco, CA, USA: Morgan Kaufmann, 1994, pp. 157–163.

[44] J. Hu and M. P. Wellman, "Multiagent reinforcement learning: Theoretical framework and an algorithm," in *Proc. 15th Int. Conf. Mach. Learn. (ICML '98)*, San Francisco, CA, USA: Morgan Kaufmann, 1998, pp. 242–250.

[45] O. Vinyals *et al.*, "StarCraft II: A new challenge for reinforcement learning," *arXiv:1708.04782 [cs]*, Aug. 2017.

[46] M. Samvelyan *et al.*, "The StarCraft multi-agent challenge," *arXiv preprint arXiv:1902.04043*, Dec. 2019. [Online]. Available: https://arxiv.org/abs/1902.04043

[47] B. Nardi and J. Harris, "Strangers and friends," in *Proc. 2006 20th Anniversary Conf. Computer Supported Cooperative Work - CSCW '06*, 2006, doi: 10.1145/1180875.1180898.

[48] B. Wu, "Hierarchical macro strategy model for MOBA game AI," in *Proc. AAAI Conf. Artif. Intell.*, vol. 33, pp. 1206–1213, Jul. 2019, doi: 10.1609/aaai.v33i01.33011206.

[49] A. K. Agogino and K. Tumer, "Unifying temporal and structural credit assignment problems," in *Proc. 17th Int. Conf. Auton. Agents Multiagent Syst. (AAMAS)*, 2004.

[50] P. Stone and M. Veloso, "Multiagent systems: A survey from a machine learning perspective," *Auton. Robots*, vol. 8, no. 3, pp. 345–383, 2000.,

[51] G. Tesauro, "Temporal difference learning and TD-Gammon," *Commun. ACM*, vol. 38, no. 3, pp. 58–68, Mar. 1995.,

[52] I. Oh, S. Rho, S. Moon, S. Son, H. Lee, and J. Chung, "Creating pro-level AI for a real-time fighting game using deep reinforcement learning," *IEEE Trans. Games*, vol. 14, no. 2, pp. 212–220, Jun. 2022.,

[53] M. Carroll *et al.*, "On the utility of learning about humans for human-AI coordination," in *Proc. 33rd Int. Conf. Neural Inf. Process. Syst. (NeurIPS)*, Dec. 2019, pp. 5174–5185.

[54] J. Wang, D. Xue, J. Zhao, W. Zhou, and H. Li, "Mastering the game of 3v3 snakes with rule-enhanced multi-agent reinforcement learning," in *Proc. 2022 IEEE Conf. Games (CoG)*, Beijing, China, 2022, pp. 229–236, doi: 10.1109/CoG51982.2022.9893608.

[55] L. Espeholt, *et al.*, "IMPALA: Scalable distributed deep-RL with importance weighted actor-learner architectures," in *Proc. 35th Int. Conf.*," *Mach. Learn.*, vol. 80, pp. 1407–1416, Jul. 10–15 2018.

[56] D. Horgan *et al.*, "Distributed prioritized experience replay," *arXiv*, Mar. 2, 2018. [Online]. Available: https://arxiv.org/abs/1803.00933.

[57] Unity Technologies, "Make a more engaging game w/ ML-Agents | Machine learning bots for game development | Reinforcement learning | Unity." [Online]. Available: https://unity.com/products/machine-learning-agents.

[58] N. Iskander, A. Simoni, E. Alonso, and M. Peter, "Reinforcement learning agents for Ubisoft's Roller Champions," *arXiv*, 2020. [Online]. Available: https://arxiv.org/abs/2012.06031.

[59] Unity Technologies, "MonoBehaviour.FixedUpdate," Unity Documentation. [Online]. Available: https://docs.unity3d.com/ScriptRef erence/MonoBehaviour.FixedUpdate.html. [Accessed: Nov. 22, 2024].

[60] M. Pleines *et al.*, "On the verge of solving Rocket League using deep reinforcement learning and sim-to-sim transfer," in *Proc. IEEE Conf. Games (CoG)*, Beijing, China, 2022, pp. 253–260, doi: 10.1109/CoG51982.2022.9893628.

[61] Y. Verhoeven and M. Preuss, "On the potential of Rocket League for driving team AI development," in *Proc. IEEE Symp. Series Comput. Intell. (SSCI)*, Canberra, ACT, Australia, 2020, pp. 2335–2342, doi: 10.1109/SSCI47803.2020.9308248.

[62] V. Moschopoulos, P. Kyriakidis, A. Lazaridis, and I. Vlahavas, "Lucy-SKG: Learning to play Rocket League efficiently using deep reinforcement learning," *arXiv*.

[63] RLGym, "RLGym: Reinforcement learning in Rocket League," [Online]. Available: https://rlgym.org/. [Accessed: Nov. 22, 2024].

[64] A. Vaswani *et al.*, "Attention is all you need," in *Proc. 31st Conf. Neural Inf. Process. Syst. (NeurIPS)*, 2017, vol. 30, pp. 6000–6010.

[65] M. Kempka, M. Wydmuch, G. Runc, J. Toczek, and W. Jaśkowski, "ViZDoom: A Doom-based AI research platform for visual reinforcement learning," in *Proc. IEEE Conf. Comput. Intell. Games (CIG)*, Santorini, Greece, 2016, pp. 1–8, doi: 10.1109/CIG.2016.7860433.

[66] M. Wydmuch, M. Kempka, and W. Jaśkowski, "ViZDoom competitions: Playing Doom from pixels," *IEEE Trans. Games*, vol. 11, no. 3, pp. 248–259, Sep. 2019.,

[67] Y. Wu and Y. Tian, "Training agent for first-person shooter game with actor-critic curriculum learning," in *Proc. Int. Conf. Learn. Representations (ICLR)*, 2016.

[68] V. Mnih *et al.*, "Asynchronous methods for deep reinforcement learning," *CoRR*, vol. abs/1602.01783, 2016. [Online]. Available: http://arxiv.org/abs/1602.01783.

[69] A. Amresh, N. Cooke, and A. Fouse, "A Minecraft-based simulated task environment for human-AI teaming," in *Proc. 23rd ACM Int. Conf. Intell. Virtual Agents*, Würzburg, Germany, 2023.

[70] K. C. Aluru, S. Tellex, J. G. Oberlin, and J. MacGlashan, "Minecraft as an experimental world for AI in robotics," in *AAAI Fall Symp.*, 2015.

[71] W. H. Guss *et al.*, "MineRL: A large-scale dataset of Minecraft demonstrations," *arXiv*, Jul. 29, 2019. [Online]. Available: https://arxiv.org/abs/1907.13440.

[72] M. Johnson, K. Hofmann, T. Hutton, and D. Bignell, "The Malmo platform for artificial intelligence experimentation," in *Proc. 25th Int. Joint Conf. Artif. Intell.*, 2016, pp. 4246–4247.

[73] "The Malmo Collaborative AI Challenge - Microsoft Research," *Microsoft Research*, Mar. 16, 2022. [Online]. Available: https://www.microsoft.com/en-us/research/academic-program/collaborat ive-ai-challenge/

[74] Y. Xiong, H. Chen, M. Zhao, and B. An, "HogRider: Champion agent of Microsoft Malmo collaborative AI challenge," in *Proc. 32nd AAAI Conf. Artif. Intell., Innovative Appl. Artif. Intell. Conf., and 8th AAAI Symp. Educ. Adv. Artif. Intell.*, 2018.

[75] D. Perez-Liebana *et al.*, "The Multi-Agent Reinforcement Learning in MalmÖ (MARLÖ) competition," *arXiv*, 2019. [Online]. Available: https://arxiv.org/abs/1901.08129.

[76] M. Jaderberg, W. M. Czarnecki, I. Dunning, L. Marris, G. Lever, A. G. Castañeda, *et al.*, "Human-level performance in 3D multiplayer games with population-based reinforcement learning," *Science*, vol. 364, no. 6443, pp. 859–865, May 31 2019.,

[77] O. Berger-Tal, J. Nathan, E. Meron, and D. Saltz, "The exploration-exploitation dilemma: A multidisciplinary framework," *PLoS One*, vol. 9, no. 4, p. e95693, Apr. 22 2014.,

[78] The AlphaStar team, "AlphaStar: Mastering the real-time strategy game StarCraft II," Google DeepMind, Jan. 24, 2019. [Online]. Available: https://deepmind.google/discover/blog/alphastar-mastering-the-real-time-strategy-game-starcraft-ii/.

[79] L. Han, *et al.*, "Grid-wise control for multi-agent reinforcement learning in video game AI," in *Proc. 36th Int. Conf.*," *Mach. Learn.*, vol. 97, pp. 2570–2585, 2019.

[80] P. Peng *et al.*, "Multiagent bidirectionally-coordinated nets: Emergence of human-level coordination in learning to play StarCraft combat games," *arXiv*, 2017. [Online]. Available: https://arxiv.org/abs/1703.10069.

[81] D. Ye, *et al.*, *Towards playing full MOBA games with deep reinforcement learning*, vol. 33. 2020, Adv. Neural Inf. Process. Syst., pp. 621–632.

[82] L. Hanke and L. Chaimowicz, "A recommender system for hero line-ups in MOBA games," in *Proc. 13th AAAI Conf. Artif. Intell. Interactive Digit. Entertain.*, 2017.

[83] Z. Chen *et al.*, "The art of drafting: A team-oriented hero recommendation system for multiplayer online battle arena games," *arXiv*, 2018. [Online]. Available: https://arxiv.org/abs/1806.10130.

[84] B. Aytemiz, M. Jacob, and S. Devlin, "Acting with style: Towards designer-centred reinforcement learning for the video games industry," in *CHI Workshop on Reinforcement Learning for Humans, Computer, and Interaction (RL4HCI)*, May 2021, p. 16.